\documentclass[lettersize,journal]{IEEEtran}
\usepackage{amsmath,amsfonts}
\usepackage{algorithmic}
\usepackage{algorithm}
\usepackage{array}
\usepackage[caption=false,font=footnotesize,labelfont=rm,textfont=rm]{subfig}
\usepackage{textcomp}
\usepackage{stfloats}
\usepackage{url}
\usepackage{verbatim}
\usepackage{graphicx}
\usepackage{cite}
\usepackage{hyperref}
\usepackage{cleveref}
\usepackage{booktabs}
\usepackage{color}
\newtheorem{proposition}{Proposition}
\newtheorem{theorem}{Theorem}
\newtheorem{proof}{Proof}[section]

\begin{document}

\title{Rethinking Domain Generalization: Discriminability and Generalizability}

\author{
Shaocong Long$^\star$, 
        Qianyu Zhou$^\star$,
        Chenhao Ying$^\dag$, 
        Lizhuang Ma,
        Yuan Luo$^\dag$
\thanks{S. Long, Q. Zhou, C. Ying, L. Ma, Y. Luo are with the Department of Computer Science and Engineering, Shanghai Jiao Tong University, Shanghai 200240, China~(email: \{longshaocong, zhouqianyu, yingchenhao, yuanluo\}@sjtu.edu.cn and ma-lz@cs.sjtu.edu.cn).
C. Ying and Y. Luo are also with Shanghai Jiao Tong University (Wuxi) Blockchain Advanced Research Center. This work was supported in part by  National Key R\&D Program of China under Grant 2022YFA1005000. 
}
\thanks{$^\star$Equal contributions, $^\dag$Corresponding authors.}
}

\markboth{IEEE TRANSACTIONS ON CIRCUITS AND SYSTEMS FOR VIDEO TECHNOLOGY,~Vol.XX, No.X, JUNE~2024}%
{Shell \MakeLowercase{\textit{et al.}}: A Sample Article Using IEEEtran.cls for IEEE Journals}

\IEEEpubid{\begin{minipage}{\textwidth}\centering \vspace{9pt}
Copyright © 2024 IEEE. Personal use of this material is permitted.\\
However, permission to use this material for any other purposes must be obtained from the IEEE by sending an email to pubs-permissions@ieee.org.
\end{minipage}}
\IEEEpubidadjcol

\maketitle

\begin{abstract}

Domain generalization~(DG) endeavours to develop robust models that possess strong generalizability while preserving excellent discriminability. Nonetheless, pivotal DG techniques tend to improve the feature generalizability by learning domain-invariant representations, inadvertently overlooking the feature discriminability. On the one hand, the simultaneous attainment of generalizability and discriminability of features presents a complex challenge, often entailing inherent contradictions. This challenge becomes particularly pronounced when domain-invariant features manifest reduced discriminability owing to the inclusion of unstable factors, \emph{i.e.,} spurious correlations. On the other hand, prevailing domain-invariant methods can be categorized as category-level alignment, susceptible to discarding indispensable features possessing substantial generalizability and narrowing intra-class variations. To surmount these obstacles, we rethink DG from a new perspective  that  concurrently imbues features with formidable discriminability and robust generalizability, and present a novel framework, namely, Discriminative Microscopic Distribution Alignment~(DMDA). DMDA incorporates two core components: Selective Channel Pruning~(SCP) and Micro-level Distribution Alignment~(MDA). Concretely, SCP attempts to curtail redundancy within neural networks, prioritizing stable attributes conducive to accurate classification. This approach alleviates the adverse effect of spurious domain-invariance and amplifies the feature discriminability. Besides, MDA accentuates micro-level alignment within each class, going beyond mere category-level alignment. This strategy accommodates sufficient generalizable features and facilitates within-class variations. Extensive experiments on four benchmark datasets corroborate that DMDA achieves comparable results to state-of-the-art methods in DG, underscoring the efficacy of our method.
The source code will be available at https://github.com/longshaocong/DMDA.
\end{abstract}


\section{Introduction}

\begin{figure}[t!]
        \centering
        \subfloat{\includegraphics[width=0.95\linewidth]{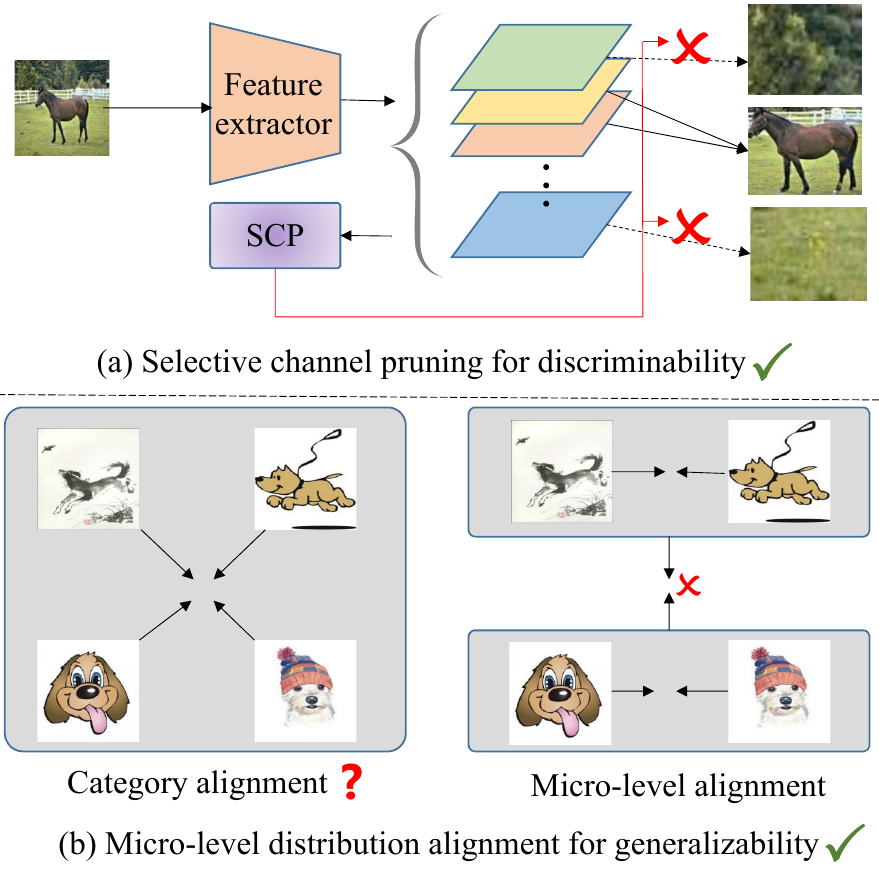}}
        \vspace{-2mm}
        \caption{
        Almost all DG methods tend to improve the feature generalizability by learning domain-invariant representations, and inadvertently overlook the feature discriminability, leading to spurious domain invariance. (a) To boost the feature discriminability, we propose Selective Channel Pruning~(SCP) to filter out unstable factors, \emph{i.e.,} spurious correlations, thereby mitigating the adverse effects of such correlations. (b) Besides, we introduce Micro-level Distribution Alignment~(MDA) to prevent the risk of discarding indispensable generalizable features in previous category-level distribution alignment. MDA could accommodate sufficient generalizable features while simultaneously enhancing within-class variations, thereby promoting feature generalizability.
        }
        \label{overview}
        \vspace{-4mm}
    \end{figure}

\IEEEPARstart{C}{omputer} vision has achieved remarkable success in  image classification~\cite{krizhevsky2012imagenet, he2016deep, dosovitskiy2020image,feng2022dmt,song2023rethinking}, image segmentation~\cite{long2015fully, ronneberger2015u, chen2018encoder}, and object detection~\cite{ren2015faster, lin2017feature, redmon2016you,zhou2022transvod,he2021end}. Nonetheless, real-world data frequently experiences distribution shifts~\cite{li2022bounds, PerKugSch22, fang2020rethinking, wang2022feature, tian2022unsupervised, zhou2022context, ren2022uni3da} across distinct scenarios, significantly impairing the performance of learned models~\cite{hendrycks2018benchmarking, taori2020measuring}. Such distribution shifts may arise from multiple factors, such as alterations in background~\cite{beery2018recognition}, changes in visual angles~\cite{scholkopf2021causal}, and camera's field of views~\cite{gu2021pit}, and \emph{etc.}. The degrading performance originates from spurious correlations captured by the model trained in limited training environments. Taking the image classification~\cite{beery2018recognition} as an instance, the hypothesis model suffers from recognizing cows on the beach due to the background shift from grassland to beach. In such case, the model may rely solely on the background instead of focusing on the presence of animals for classification. As such, the model's ability to generalize to out-of-sample scenarios is compromised.
\IEEEpubidadjcol

Numerous efforts have been dedicated to enhancing models' generalization capacities~\cite{fu2019geometry, scholkopf2021causal, pearl2013probabilistica, wei2022multi, meng2022attention}, with the purpose of capturing genuine correlations from biased data. 
Unsupervised domain adaptation~(UDA)~\cite{ganin2016domainadversarial, zhang2020domain, tian2022unsupervised, zuo2021margin,zhou2022context,zhou2022generative,zhou2022uncertainty,zhou2022domain,zhou2023self} stands as an effective technique to mitigate distribution shifts. Nonetheless, these models require access to the target data during the training phase and thereby necessitate adaption when new scenarios occur, which is time-consuming and impractical in real-world scenarios. A more challenging avenue to address distribution shift is domain generalization~(DG)~\cite{muandet2013domain,li2018domain, zhao2020domain,zhou2023instance, zhao2022style, li2023sparse}, which has no access to the target data during training and exclusively leverages information from source domains.

\begin{figure}[t!]
        \centering
        \subfloat[Target: Cartoon]{\includegraphics[width=1.7in]{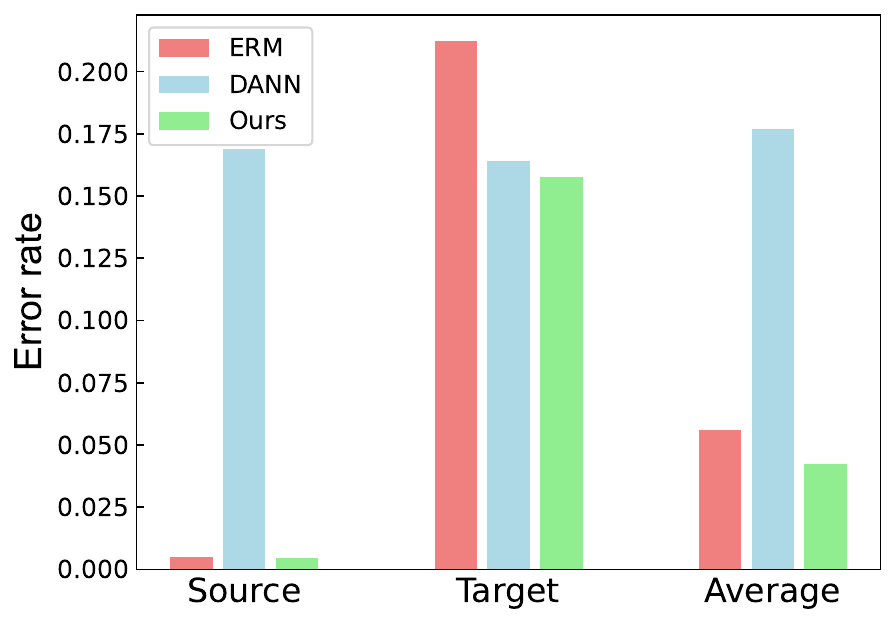}}
        \subfloat[Target: Sketch]{\includegraphics[width=1.7in]{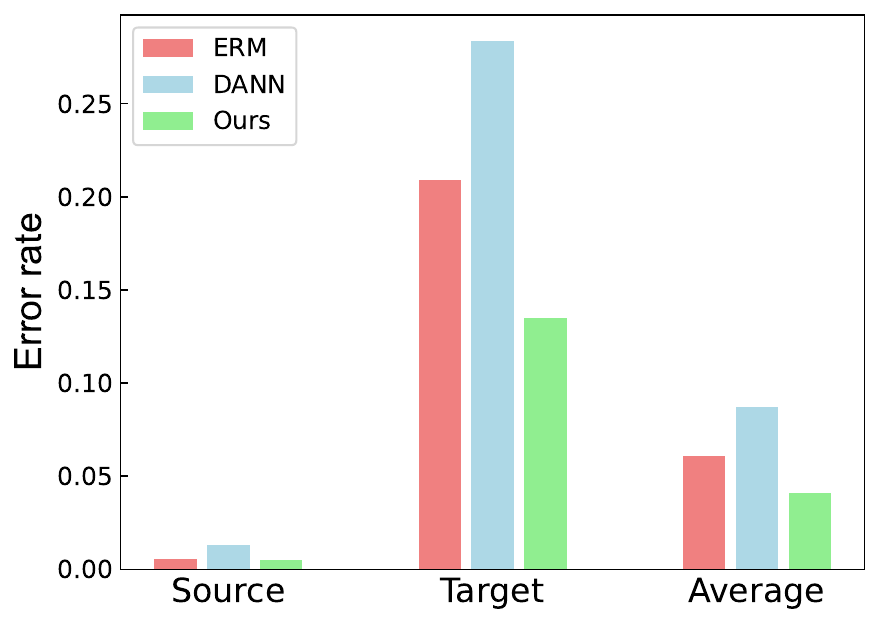}}
       \caption{Comparisons of classification error rates on the feature representations. The error rate measures the disciriminability of acquired features.}
        \label{classification_error}
        \vspace{-5mm}
    \end{figure}

Current approaches in DG tend to improve the feature generalizability via learning domain-invariant representations~\cite{motiian2017unified, li2018domain_2, sun2016deep, zhao2020domain, ganin2016domainadversarial} while overlooking and compromising the feature discriminability. It is noteworthy that there is an inconsistency in simultaneously improving the generalizability and discriminability of features. Fig.~\ref{classification_error} measures the feature discriminability by measuring the classification error rate with the acquired features. As observed, the error rate on the representation of DANN~\cite{ganin2016domainadversarial} exceeds that of ERM in most scenarios, indicating the reduced feature discriminability by DANN. Consequently, it is significant to mitigate the mutual influence between generalizability and discriminability. However, it is non-trivial to design an effective DG mechanism to trade off the generalizability and discriminability that is applicable in various scenarios. On the one hand, excessive emphasis on feature discriminability compromises the generalizability. On the other hand, excessive focus on generalizability poses the peril of generating features deficient in discrimination. We argue that the crux lies in the unstable factors, \emph{i.e.}, spurious correlations. When acquired features lack stable relevance to classification, pursuing domain-invariance on such features becomes futile and may even intensify source risks. Hence, it is crucial to enforce domain-invariance upon stable factors, which can mitigate the adverse effect of unstable factors on the consistency between generalizability and discriminability.

In addition to addressing the challenges posed by unstable factors, it is essential to acknowledge that achieving domain-invariant learning can be an intricate endeavor. Conventional domain-invariant learning is characterized as category alignment, wherein the predominant approach seeks to mitigate disparities amongst prototypes or semantics of samples within each category across domains~\cite{li2020simultaneous, zhao2022semantic, motiian2017unified, yao2022pcl, li2018domain, zhao2020domain}. However, such category alignment inherently is a coarse-grained alignment strategy and may confront difficulties in learning sufficient generalizable features. This is chiefly attributed to the relative substantial divergence in attributes for samples within the same category, rendering the coarse-grained alignment for a specific category across domains potentially liable to discard indispensable features possessing substantial generalizability. As illustrated in Fig.~\ref{overview}(b), there is a significant disparity between running dogs and dogs depicted with only their heads. Attempting to align them forcibly could result in the loss of distinctive features associated with the dog's body and legs. Therefore, it is significant to relax the stringent constraints of category-level domain-invariance. In doing so, we focus on aligning the distributions of samples within a category that have similar semantics, without mandating alignment for samples with distant semantics within the same category. This way also serves to enrich the within-class variations, a known catalyst for enhancing generalization performance\cite{zhang2022principled, mahajan2021domain}.

To surmount the aforementioned limitations, we rethink DG through the lens of feature discriminability and generalizability. In this work, we present a novel framework, named Discriminative Microscopic Distribution Alignment~(DMDA). This framework comprises two innovative modules: Selective Channel Pruning~(SCP) and Micro-level Distribution Alignment~(MDA), which collaborate to elevate the feature discriminability and generalizability. Fig.~\ref{overview} illustrates the two key modules in our proposed approach. Specifically, SCP filters out the unstable channels in features and curtails redundancy within neural networks to enhance feature discriminability. 
Besides, MDA performs distribution alignment at the micro level rather than the category level to promote feature generalizability. Compared with existing domain-invariant methods in DG, our approach offers dual advantages: Firstly, our model focuses on stable correlations, aiming to heighten the feature discriminability. Secondly, our strategy guarantees micro-level invariance, facilitating the acquisition of sufficient generalizable features and accommodating within-class variations. Consequently, our proposed approach not only augments feature generalizability across domains but also enhances category-level discriminability, resulting in superior generalization capacities of models.
In summary, our main contributions include:
\begin{itemize}
    \item We introduce a novel perspective for DG with the dual objectives of enhancing feature generalizability while concurrently improving discriminability.
    \item We present an innovative approach named DMDA for DG, comprising two key modules: SCP and MDA. Concretely, SCP mitigates the detrimental effect of spurious correlations, thereby endowing the acquired features with heightened discriminability. While MDA pursues domain invariance at the micro level, rather than the category level, boosting feature generalizability.
    \item Extensive experiments on four benchmark datasets corroborate the efficacy and superiority of the proposed approach, where it achieves competitive performance compared to state-of-the-art methods in DG.
\end{itemize}

\section{Related Work}

Techniques from diverse perspectives have been proposed to improve models' generalization capacity in DG. 

One notable research avenue is domain-invariant learning~\cite{ghifary2016scatter, muandet2013domain,motiian2017unified, gao2022multi}. The objective is to cultivate common features across domains that possess the potential to generalize effectively to novel scenarios. Denote the input, acquired features, and output as $X$, $\Phi$, and $Y$, respectively. A foundational approach, proposed by Ganin~\emph{et al.}~\cite{ganin2016domainadversarial} as Domain Adversarial Training of Neural Networks (DANN), seeks to acquire domain-invariant features through adversarial learning, under the assumption that the conditional distribution $P(Y|X)$ remains invariant. However, this assumption may not hold in practical settings, leading to DANN's limitations. To address this limitation, Li~\emph{et al.}~\cite{li2018deep} proposed a conditional adversarial network to satisfy the invariant distribution $P(\Phi|Y)$, considering the causal direction $Y \rightarrow X$ and assuming class-balance in the target data. Alternatively, Zhao~\emph{et al.}~\cite{zhao2020domain} proposed minimizing the divergence between conditional distributions across domains, ensuring the invariant conditional distribution $P(Y|\Phi)$. Additionally, self-supervised learning~\cite{grill2020bootstrap, he2020momentum, chen2021exploring, zhao2022embedding, chen2023instance} has made notable contributions to improve the expressive power of features. SelfReg~\cite{kim2021selfreg} and PCL~\cite{yao2022pcl}, for instance, took advantage of contrastive learning to encourage representations to be close for samples within the same class and distant from samples in other classes. Besides, a heightened awareness has emerged regarding the robustness of feature maps~\cite{du2022cross, guo2023domaindrop}. These methods are directed towards suppressing domain-sensitive information from a feature map perspective. 

Another intuitive approach is to expand the source domains through data augmentation 
\cite{shankar2018generalizing, xu2021fourier, wang2022feature, li2023exploring} or data generation~\cite{zhou2020learning}, thus indirectly decreasing the distribution gap between source and target domains. 
For instance, Zhou~\emph{et al.}~\cite{zhou2020deep} proposed a transformation for data augmentation using adversarial training. Rahman~\emph{et al.}~\cite{rahman2019multi} employed ComboGAN~\cite{anoosheh2018combogan} to generate new data with a small distribution discrepancy to the original data. 

Ensembling techniques~\cite{cha2021swad, kim2021selfreg, chu2022dna} have been introduced to seek flatter minima, which has been proven an effective strategy for enhancing generalization performance. SelfReg~\cite{kim2021selfreg} and SWAD~\cite{cha2021swad} utilized Stochastic Weights Averaging to find the flatter minima, consequently reducing the domain generalization gap in the target domain.

Different from prevailing approaches primarily focused on enhancing the generalizability of acquired features in DG, this study endeavors to concurrently elevate both feature discriminability and generalizability.
\begin{figure*}[t!]
        \centering
        \captionsetup[subfigure]{oneside,margin={0.5cm,0cm}}
        \subfloat[Target: Art]{\includegraphics[width=1.7in]{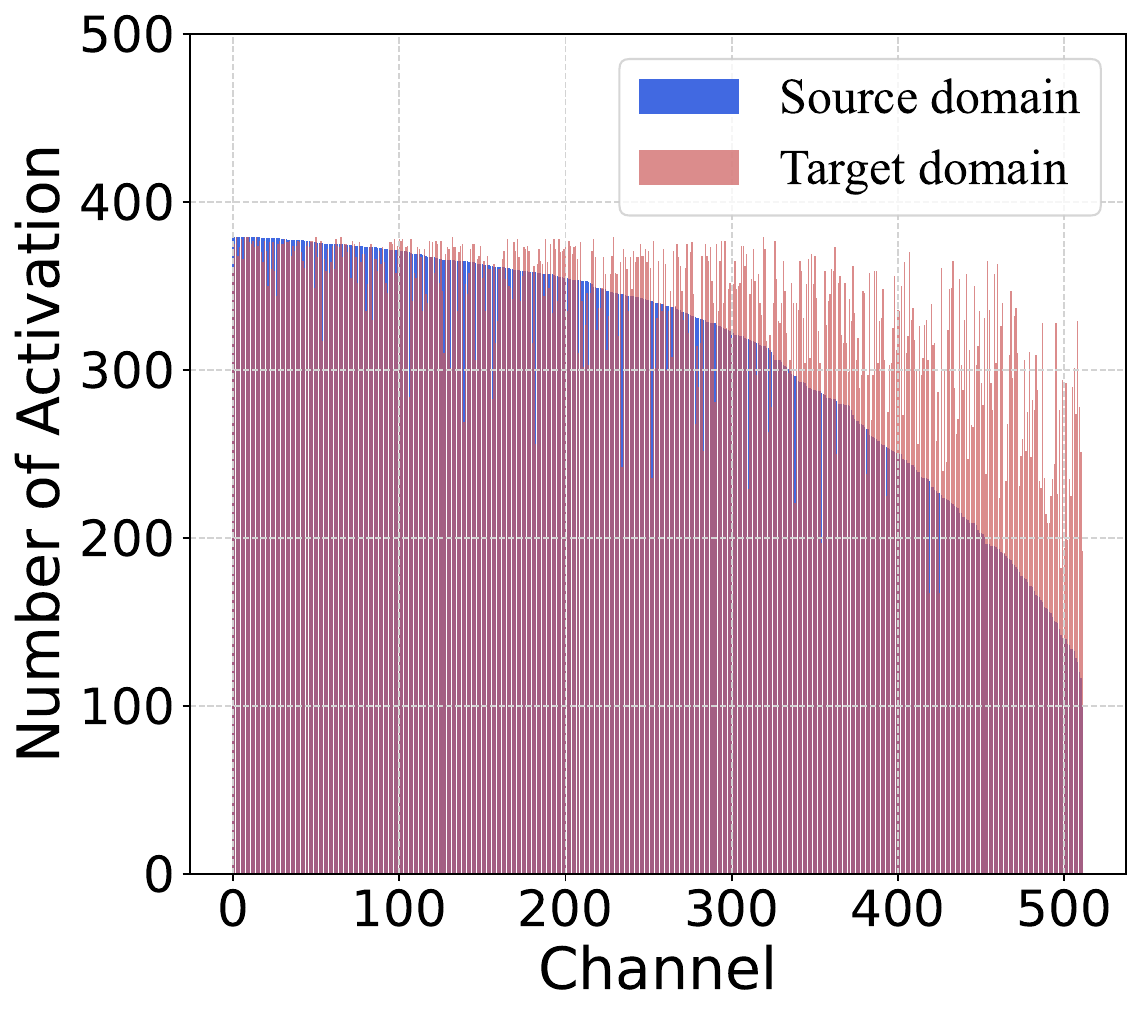}}
         \hspace{3pt}
         \captionsetup[subfigure]{oneside,margin={0.5cm,0cm}}
        \subfloat[Target: Cartoon]{\includegraphics[width=1.7in]{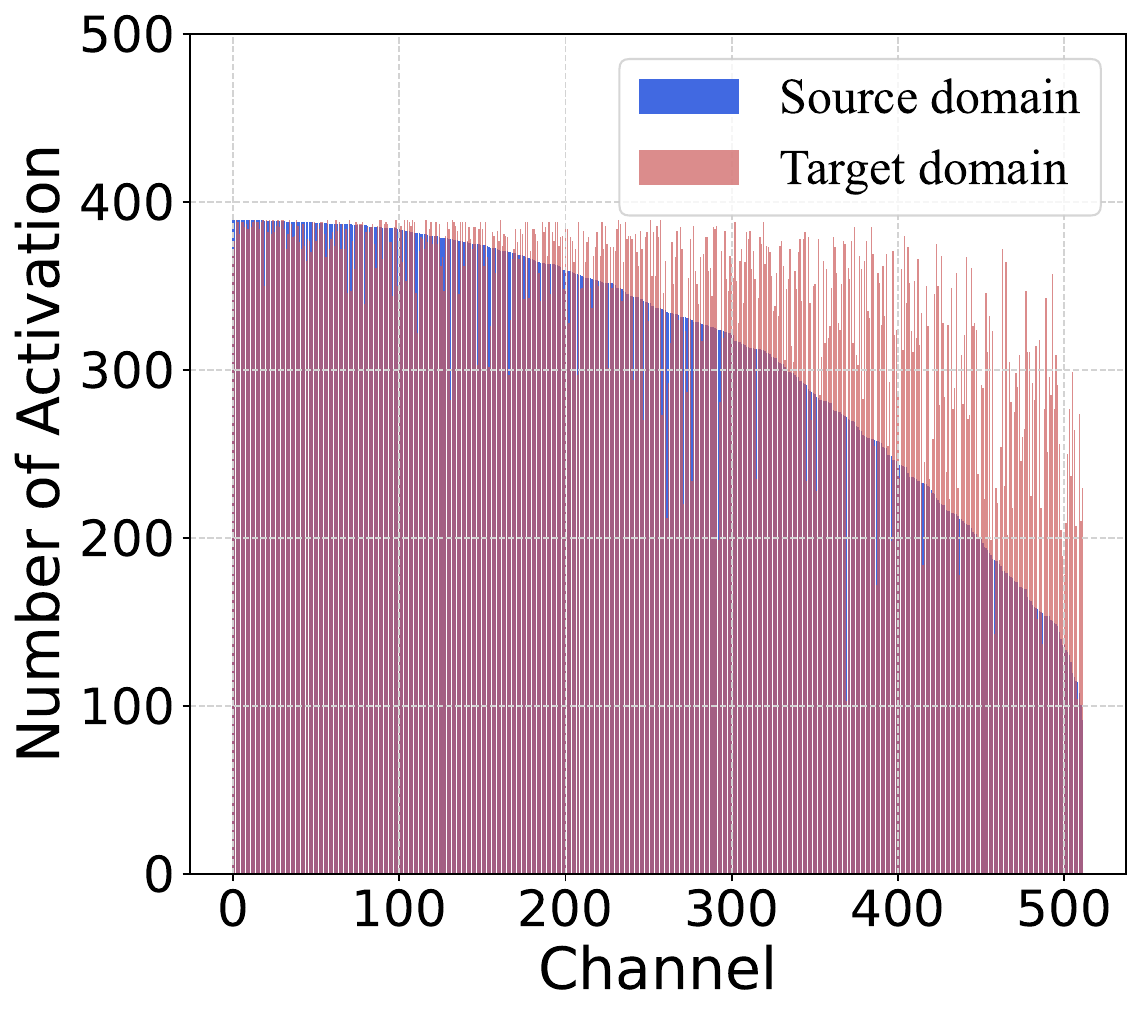}}
         \hspace{3pt}
         \captionsetup[subfigure]{oneside,margin={0.5cm,0cm}}
        \subfloat[Target: Photo]{\includegraphics[width=1.7in]{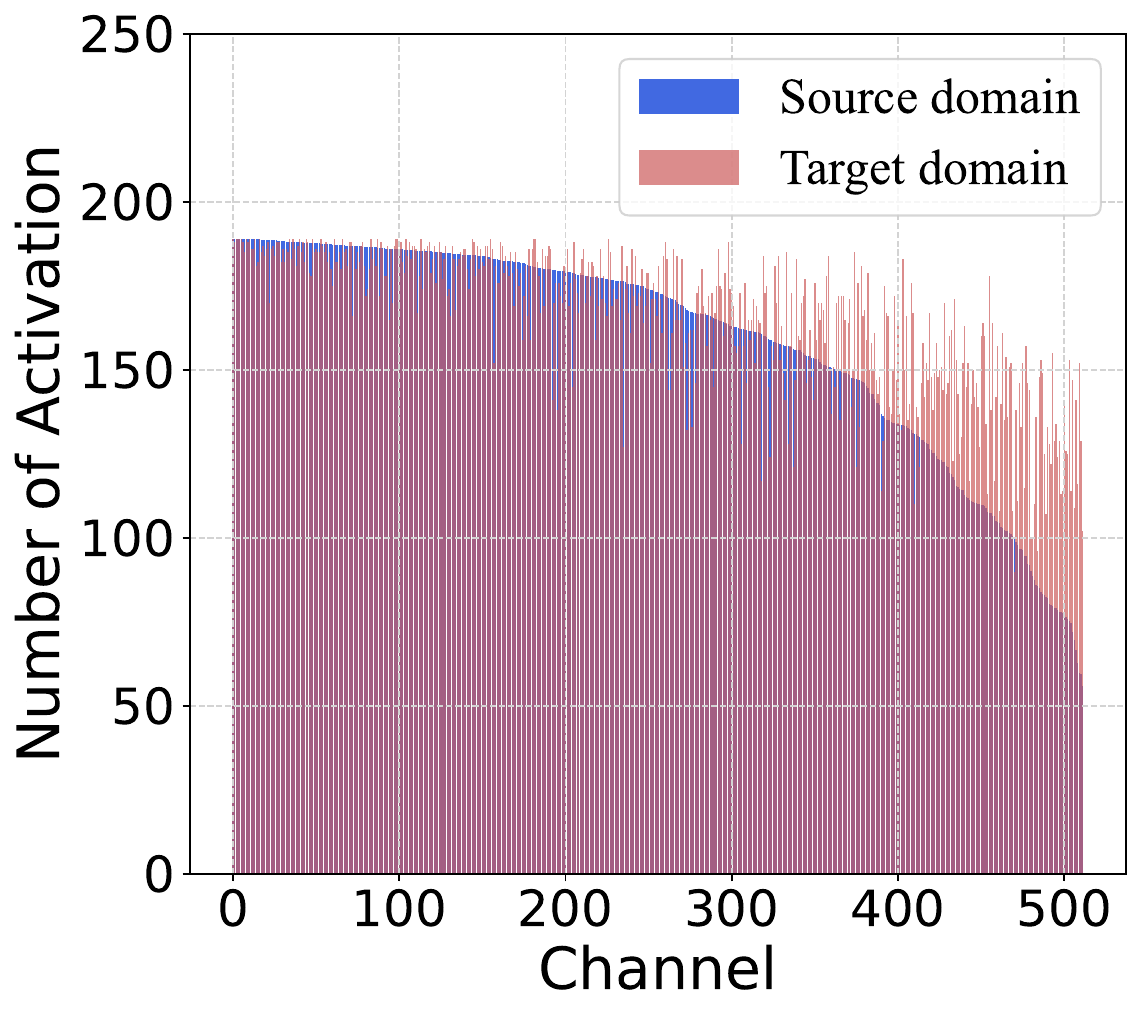}}
        \hspace{3pt}
        \captionsetup[subfigure]{oneside,margin={0.5cm,0cm}}
        \subfloat[Target: Sketch]{\includegraphics[width=1.7in]{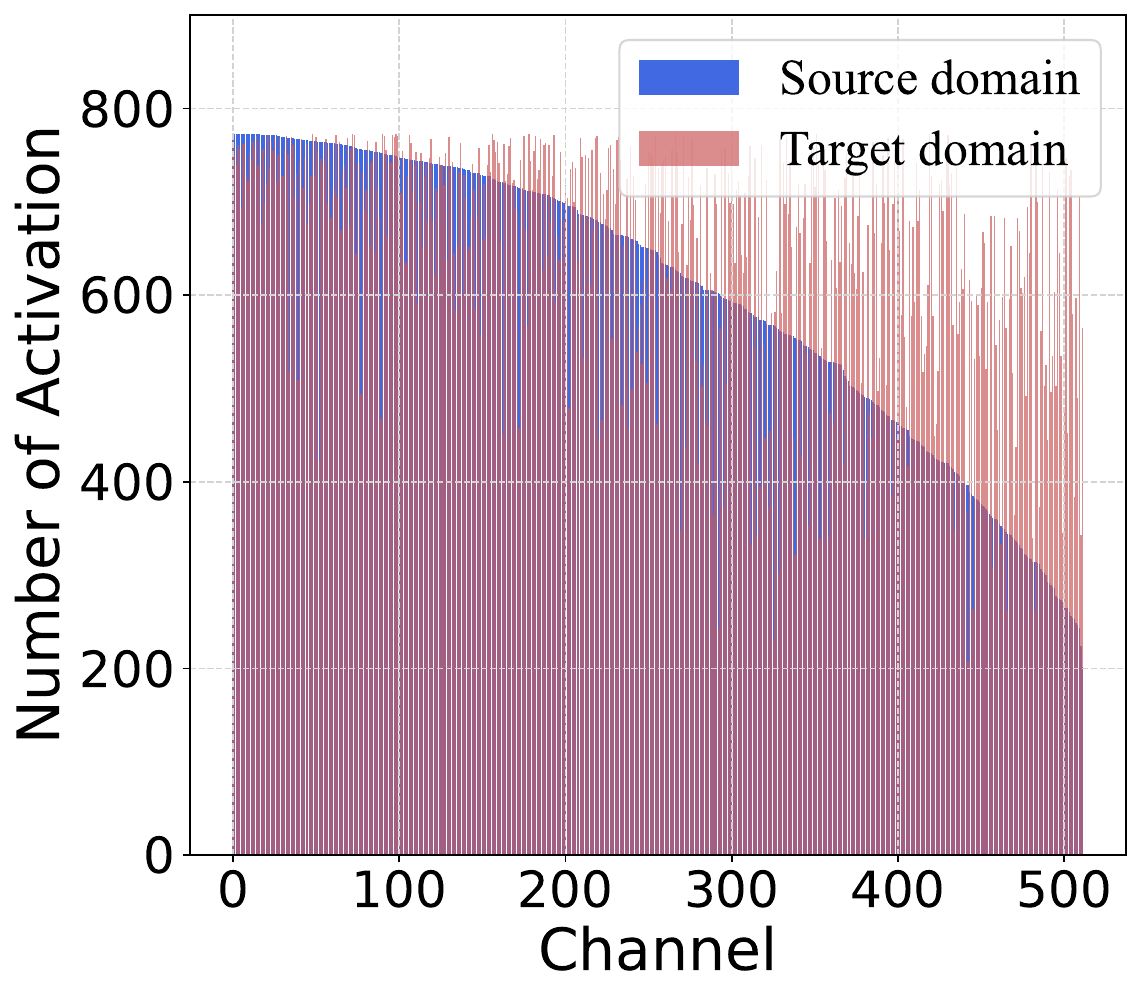}}
        \centering
        \caption{The channel activation frequency in the penultimate layer of ResNet-18 trained via ERM, with `Art', `Cartoon', `Photo', and `Sketch' as the target domain, respectively. Channels are arranged in descending order based on the activation frequency in source domains. Channels experiencing infrequent activation in source domains but displaying frequent activation in target domain can be characterized as unstable factors demonstrating spurious correlations.}
        \label{ERM_activation}
    \vspace{-1em}
    \end{figure*}

\section{Methodology}
\label{method}

\subsection{Preliminaries}
Let $\mathcal{X}$ and $\mathcal{Y}$ denote the input space and output space, respectively. In DG, there are $M$ source (seen) domains $S_{source} = \{S^i | i= 1, 2, \cdots, M\}$ 
where $S^i = \{(x_j^i, y_j^i)|j = 1, 2, \cdots, n_i\} \sim P_i^S(X, Y)$. Here, $n_i$ is the number of samples in domain $S^i$, and $P_i^S(X, Y)$ is the joint distribution of the covariates together with labels in domain $S^i$, which is different from that of other domains: $P_i^S(X, Y) \neq P_j^S(X, Y), i \neq j$. We assign $U_{source} = \{U^i | i= 1, 2, \cdots, M\}$ as the domain-dependent variables (style, background, \emph{etc.}), which vary across domains: $U^i \neq U^j, i \neq j$. Given the source domains, the goal of DG is to learn a robust model $h = g \circ f$, where $f : X \rightarrow \Phi$ is the representation function and $g : \Phi \rightarrow Y$ is the label predictive function. This model is expected to generalize well to the $N$ target (unseen) domains $T_{target} = \{T^i | i= 1, 2, \cdots, N\}$ 
where $T^i = \{(x_j^i, y_j^i)|j = 1, 2, \cdots, n_i\} \sim P_i^T(X, Y)$. The distributions in source domains and target domains are different: $P^S(X, Y) \neq P^T(X, Y)$. Besides, the target domains cannot be accessed during the training phase.

\begin{figure*}[t!]
    \centering
    \includegraphics[width=.95\linewidth]{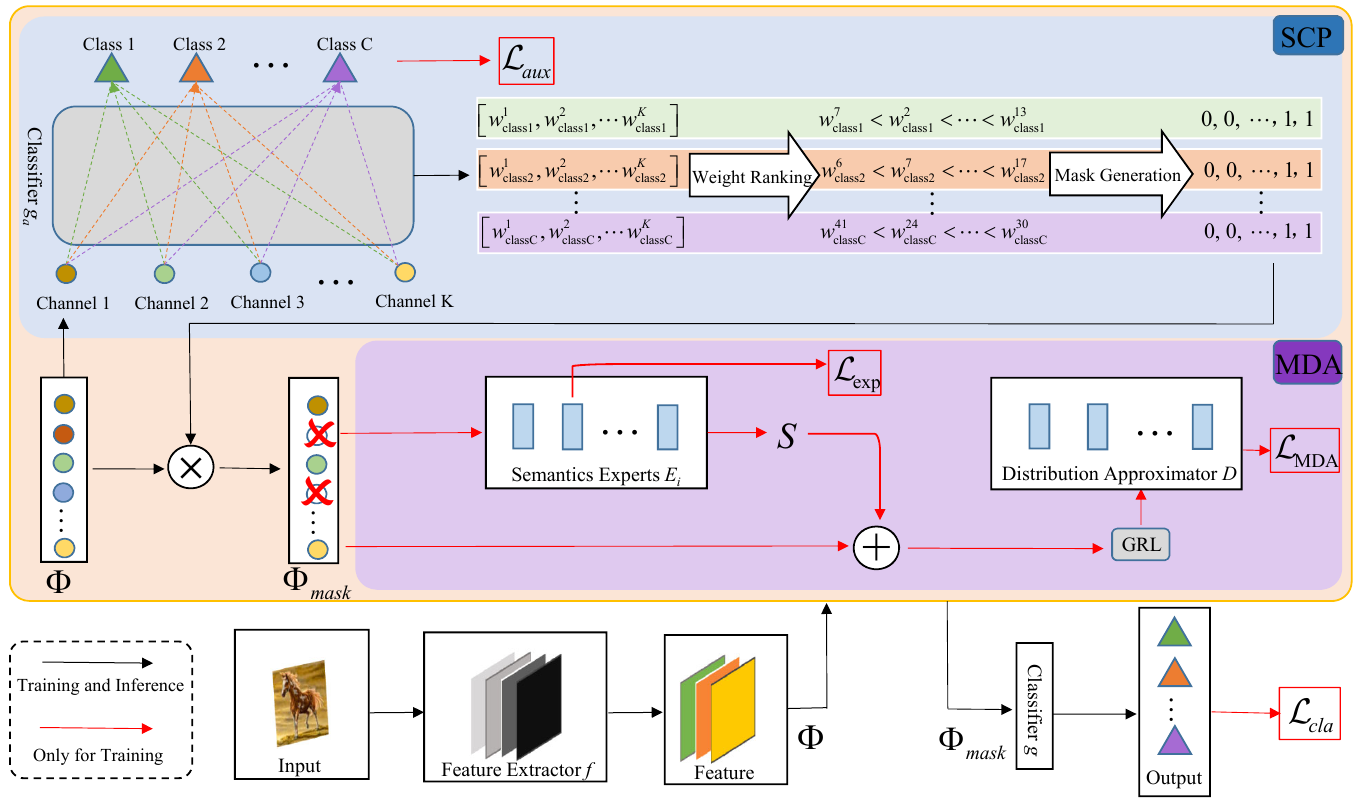}
    \caption{\textbf{Framework of our proposed Discriminative Microscopic Distribution Alignment~(DMDA).}
    The features generated by the feature extractor are initially transmitted to the Selective Channel Pruning (SCP) module to create channel-wise masks for the purpose of eliminating unstable channels. Following this, the features undergo pruning based on the channel-wise masks. Subsequently, Micro-level Distribution Alignment (MDA) is employed to execute micro-level distribution alignment rooted in the latent semantics of the pruned features, which are generated by additional semantics experts. \label{framework}}
   
    \vspace{-1em}
\end{figure*}

In this section, we elucidate our strategy for overcoming the inherent constraints of domain-invariant methods in DG, endowing the acquired features with both robust generalizability and formidable discriminability.  Fig.~\ref{framework} illustrates the overall architecture of our proposed Discriminative Microscopic Distribution Alignment~(DMDA), encompassing two key modules, namely Selective Channel Pruning~(SCP) and Micro-level Distribution Alignment~(MDA). SCP serves to ameliorate the impact of unstable factors on the features discriminability, while MDA augments features' generalizability through micro-level distribution alignment. Specifically, the features acquired through the feature extractor traverse through SCP to derive channel-wise masks, which are subsequently employed for feature pruning. Subsequently, MDA aligns micro-level distributions of the pruned features, guided by the underlying semantics of the pruned features.

\subsection{Selective Channel Pruning}
Prevalent DG methods strive to enhance feature generalizability by acquiring domain-invariant representations, often inadvertently sidelining discriminability. Nevertheless, the relentless pursuit of domain invariance can yield features with subpar discriminability, posing a formidable challenge for the classification. As illustrated in Fig.~\ref{classification_error}, the drive for domain invariance by DANN\cite{ganin2016domainadversarial} can lead to diminished feature discriminability. This dilemma arises from the conflict between the concurrent enhancement of both generalizability and discriminability. We posit that this may result from unstable factors, \emph{i.e.,} spurious correlations, which induce spurious domain invariance while concurrently impeding discriminability.

To succeed in a classification task, it is imperative that the acquired features encapsulate only the stable factors carrying essential information while discarding spurious correlations. The current solution typically seeks to minimize the empirical risk across source domains:
\begin{equation}
    \mathcal{L}_{cla} =\sum_{i = 1}^{M}\sum_{x, y}{P^S_i(x, y)}  \ell(h(x), y),
\end{equation}
where $\ell$ denotes the cross-entropy loss. However, the straightforward approach cannot guarantee the stability of every dimension in the acquired features. This implies that certain dimensions may convey spurious correlations for classification and consequently offer limited utility for generalization. 

Based on the above insight, we explore the stability of different channels to enhance generalization performance. When considering a specific class, channels carrying genuine information tend to manifest more generalized patterns and should be activated with greater frequency. Conversely, channels capturing spurious correlations should experience less frequent activation due to the distribution shift across source domains. To gain a comprehensive understanding of this phenomenon, we visualize the channel activation frequency within source domains and target domains in Fig.~\ref{ERM_activation}, respectively. A channel is considered activated if its activation value exceeds a threshold~(1\% of the highest activation value across all channels in our case.). In this visualization, we take the class `dog' as an instance, and analyze the activation frequency of each channel for samples in source domains and target domains, respectively. We arrange the channels in descending order based on the activation frequency of samples in source domains. It is noteworthy that we have adjusted the number of samples in the target domain to ensure a fair comparison. As depicted in Fig.~\ref{ERM_activation}, samples from source domains and target domains exhibit different activation patterns. Samples in target domains frequently activate channels that receive less frequent activation from samples in source domains. This trend persists across all classes. The channels that are less frequently activated in source domains and display inconsistencies in activation frequency between the source and target domains can be identified as unstable channels, containing spurious correlations that are detrimental to generalization. 

The above observation motivates us to propose an unstable factor removal strategy, Selective Channel Pruning~(SCP), to mitigate the adverse effects of unstable factors on generalization capacities. Denote the output of $l$-th activation layer of the model $h$ as $\Phi^l \in \mathbb{R}^{H\times W\times K}$, where $H$, $W$, and $K$ represent the height, width, and number of channels, respectively. In the SCP module, we begin by applying the Global Average Pooling~(GAP) operation on the raw feature map, resulting in the channel-wise activation $\hat{\Phi}^l \in \mathbb{R}^K$. Mathematically, the activation of the $k$-th channel can be defined as:
\begin{equation}
    \hat{\Phi}^l = \frac{1}{H \times W}\sum_{i = 1}^H \sum_{j = 1}^W \Phi_k^l(i,j).
\end{equation}
We then input the channel-wise activation $\hat{\Phi}^l$ into an auxiliary classifier $g_{a}$, which can be optimized by minimizing the risk:
\begin{equation}
    \mathcal{L}_{aux} =\sum_{i = 1}^{M}\sum_{x, y}{P_i^S(x, y)}\ell(g_{a}(\hat{\Phi}^l), y).
\end{equation}
This classifier consists of only one fully connected~(FC) layer to perform classification. Considering a classification task with $C$ classes, we denote the parameters of this auxiliary classifier as $W_l = [W_1^l, W_2^l. \cdots, W_C^l] \in \mathbb{R}^{K \times C}$, which assesses the significance of each channel to a specific class. To mitigate the adverse effects of unstable channels, we introduce a pruning strategy that takes into account the channels' significance in the classification. More specifically, the pruning strategy for $i$-th channel of class $c$ can be mathematically formulated as:
\begin{equation}
    M_{c, i}^l = \left\{
    \begin{aligned}
    0&, & W_{c, i}^l \leq Q_q(W_{c}^l)\\
    1 &, & W_{c, i}^l > Q_q(W_{c}^l)
    \end{aligned}
    \right.,
\end{equation}
where $Q_q$ represents $q$-th percentile for $W_{c}^l$, serving as the criterion for identifying whether channels are selected as stable channels. Subsequently, we employ these masks to reconstruct the original feature maps in a channel-wise manner. During the training stage, we adopt the ground-truth label $y$ as the reference to determine channel importance and generate channel masks. However, during the testing stage when access to the ground-truth label $y$ is unavailable, we employ the index~(denoted as $\overline{y}$) associated with the maximum value in the predicted labels to determine the channel importance and the channel mask selection. The resultant masked channels can be represented as:
\begin{equation}
    \Phi^l_{mask}=  \left\{
    \begin{aligned}
    &\Phi^l \otimes M_y^l, &\text{training phase}\\
    &\Phi^l \otimes M_{\overline{y}}^l, &\text{test phase}\\
    \end{aligned}
    \right.,
\end{equation}
where $\otimes$ denotes channel-wise multiplication. The pruned feature maps rather than the original feature maps will be propagated into the subsequent layer.

\subsection{Micro-level Distribution Alignment}

Methods based on domain-invariant representation learning have garnered significant attention in DG. 
Given the assumption that the conditional distribution $P(Y|X)$ stays invariant, Ganin~\emph{et al.}~\cite{ganin2016domainadversarial} proposed domain adversarial training of neural networks~(DANN) for domain adaptation, which aims to cultivate domain-invariant features $\Phi$ through adversarial learning. To address the limitation posed by assuming invariant conditional distributions, Li~\emph{et al.}~\cite{li2018deep} presented a conditional adversarial network to satisfy the invariant distribution $P(\Phi|Y)$, while considering the causal direction $Y \rightarrow X$ and assuming class-balance in the target data. As an alternative and effective strategy, Zhao~\emph{et al.}~\cite{zhao2020domain} proposed minimizing the divergence between conditional distributions across domains to ensure the invariant conditional distribution $P(Y|\Phi)$.

While these domain-invariant methods have made strides by aligning the marginal and conditional distributions, they overlook a critical aspect: As category alignment, these methods inherently are coarse-grained alignment strategies. The coarse-grained alignment is a rather strict constraint and may result in information loss within categories. Consequently, category alignment methods encounter challenges in acquiring an adequate pool of generalizable features. This challenge primarily arises due to significant variations in attributes among samples within the same category. Such variations make coarse-grained alignment strategies for specific categories across domains prone to the inadvertent exclusion of essential features that hold significant generalizability.

To mitigate the adverse effects linked to category alignment in DG, we introduce Micro-level Distribution Alignment~(MDA). MDA is designed to establish domain invariance at the micro level, a departure from the conventional category-level domain invariance. This shift alleviates the rigorous constraints associated with category-level domain invariance by concentrating on achieving invariance specifically among samples sharing similar semantics, eschewing unnecessary alignment of samples in one category with distant semantics. In practice, we pursue micro-level domain invariance by aligning features and their latent semantics across domains:
\begin{equation}
    P_1^S(\Phi, S) = P_2^S(\Phi, S) = \cdots = P_M^S(\Phi, S), \label{matching}
\end{equation}
where $S$ denotes the latent semantics of features, $(\Phi, S)$ represents the joint variable of $\Phi$ and $S$, and $P_i^S$ refers to the corresponding distribution in source domain $i$. 

The necessity for the latent semantics of features to possess the capability to discern subtle distinctions among samples within a single category is paramount for the success of micro-level alignment. In the absence of such discernment, deficient semantics may offer little assistance and, in certain cases, could even be detrimental to the alignment process, particularly when two distinct images share similar semantics. To address this challenge, we introduce a set of $M$ additional specialist experts $\{E_i\}_{i = 1}^M$, to model the latent feature semantics and thereby capture these subtle differences of samples in a certain category, each expert tailored to a specific source domain. In practice, we opt for a simplified approach, employing two fully connected layers for each expert to avoid the intricacies of more complex models. The dimension of the first layer output corresponds to the number of classes, enabling the optimization of semantic experts, while the dimension of the subsequent layer output matches that of the features. The optimization is achieved through the minimization of cross-entropy loss between the outputs in the first layer, denoted as $E_i^1(\Phi)$, and the ground truth:
\begin{equation}
    \mathcal{L}_{exp} = \sum_{i = 1}^{M}\sum_{x, y}{P_i^S(x, y)}\ell(E_i^1(\Phi), y).
\end{equation}
In practical application, we utilize experts' outputs, denoted as $S = E_i(\Phi)$, as surrogates for latent semantics. The first layer outputs of experts preserve predictive scores of other categories, in contrast to ground-truth labels which omit information of other categories. The predicted class scores may exhibit variations even for samples within the same category. As a consequence, the predicted latent semantics $S$, based on the first layer outputs, is well-suited to capturing nuanced distinctions among samples in one category. Consequently, they effectively meet the criteria necessary for feature latent semantics to facilitate micro-level alignment.

\noindent\textbf{Micro-level Distribution Alignment vs. Category Alignment.} It is of paramount importance to establish micro-level invariance based on experts' predicting semantics rather than enforcing category-level invariance solely relying on ground-truth labels across source domains. This necessity arises from the fact that attributes of different images within the same category may exhibit significant variations. Consequently, aligning class-to-class samples may emphasize invariant yet less significant features, potentially having an adverse effect on the acquisition of features with high generalizability and subsequently increasing empirical risk. In this way, category alignment may offer limited benefits and compromise classification accuracy. In contrast, the outputs of these experts encompass a spectrum of latent semantics across source domains, capable of reflecting various sample attributes in each class. Hence, these predicting outputs can be considered as the latent feature semantics for micro-level alignment. Consequently, MDA exhibits increased robustness to samples with significant variations and is more tolerant with within-class variations, thereby enhancing generalization performance.

In practice, we lack direct access to the semantic-based distributions 
$\{P_1^S(\Phi, S), P_2^S(\Phi, S), \cdots, P_M^S(\Phi, S)\}$. This absence of access hinders our ability to match distributions across source domains. To tackle this challenge, we employ mutual information as a measure to assess the independence between the domain-dependent variables $U$ and the semantic-based variables $(\Phi, S)$. Consequently, we optimize the feature extractor by minimizing the mutual information:
\begin{equation}
\begin{split}
    &\arg \min_{f} I(U; (\Phi, S))\\
    = &\arg \min_{f}\{ H(U) - H(U|(\Phi, S))\}\\
    = &\arg \min_{f}- H(U|(\Phi, S))\\
    =& \arg \min_{f} \mathbb{E}_{P^S(X, Y, U)}[\log P(U|(\Phi, S))] \\
    = &\arg \min_{f} \mathbb{E}_{P^S(X, Y)}[\log P(U|(\Phi, S))] \label{mutual},
\end{split}
\end{equation}
where $P(U|(\Phi, S))$ represents the probability of $U$ conditioned on the semantic-based variables $(\Phi, S)$. It is worth noting that as domain-dependent variable $U$ remains constant throughout the feature extractor optimization process, the first term $H(U)$ can be safely omitted from the objective. However, the unavailability of the distribution $P(U|(\Phi, S))$ still poses a challenge for minimization. 

To facilitate the minimization, we introduce a distribution approximator $D$ designed to approximate the conditional distribution as: $D(\Phi, S) := P(U|(\Phi, S))$. Given the intractability of the conditional distribution, we are unable to directly evaluate the quality of the function $D$. As a consequence, if the predictions of $D$ are inaccurate, the mutual information minimization in Eq.~\eqref{mutual} based on such predictions makes nonsense. In response, we reframe the minimization of the negative conditional entropy $- H(U|(\Phi, S))$ in Eq.~\eqref{mutual} as a more challenging problem: optimizing the worst risk of mutual information by minimizing the supremum of the approximated negative conditional entropy, where we denote $\mathbb{E}_{P^S(X, Y)}\log D(\Phi, S)$ as $-H(D(\Phi, S))$:
\begin{equation}
\begin{split}
    \min_{f} \sup_{D} \mathbb{E}_{P^S(X, Y)}\log D(\Phi, S)                       \label{mutual_sup}.
\end{split}
\end{equation}
Turning the intractable negative entropy $- H(U|(\Phi, S))$ into the supremum of the approximated negative conditional entropy $-H(D(\Phi, S))$ makes the optimization of Eq.~\eqref{mutual} achievable. Considering that the range of $D$ is closed, we employ the following minimax game to address Eq.~\eqref{mutual_sup}:
\begin{equation}
\begin{split}
  &\min_{f} \max_{D} \mathbb{E}_{P^S(X, Y)}\log D(\Phi, S)\\ = 
    & \min_{f} \max_{D} \sum_{i = 1}^{M}\sum_{x,y}{P^S_i(x, y)}\log D_i(\Phi, S)\\=
    & \min_{f} \max_{D} \sum_{i = 1}^{M}\sum_{x,y}{P^S_i(x, y)}\log D_i(\Phi, E_i(\Phi)). \label{minmax}
\end{split}
\end{equation}%

\begin{algorithm}[tb]
\caption{Training algorithm for DG via DMDA}
\label{algorithm}
\textbf{Input}: $M$ source training datasets: $\{S_i\}_{i = 1}^{M}$\\
\textbf{Parameter}: Weighting factor: $\alpha$, $\beta$, quantile parameter: $m$ \\
\textbf{Output}: Feature extractor: $f$, semantics experts: $\{E_i\}_{i = 1}^M$, classifiers: $g$, $g_a$, distribution approximator: $D$\\
\vspace{-0.5cm}
\begin{algorithmic}[1] 
\WHILE{training is not converged}
\FOR{$i = 1$ to $M$}
\STATE Sample data from $S_i$
\STATE Generate the channel mask and prune the features 
\STATE Calculate the latent semantics of pruned features
\STATE Sum the pruned features and the predicted latent semantics to surrogate the joint distributions
\STATE Update $D$ by maximizing Eq.~\eqref{all_loss}
\STATE Update $f$, $g$, $g_a$, and $E_i$ by minimizing Eq.~\eqref{all_loss}
\ENDFOR
\ENDWHILE
\end{algorithmic}
\end{algorithm}
The subsequent proposition and theory elucidate that the minimax game in Eq.~\eqref{minmax} is equivalent to aligning the semantic-based distributions $P(\Phi, S)$ across source domains.
\begin{proposition}
Let $\Phi^{\prime} = f(X)$ for a fixed representation function $f$ and $S^{\prime} = \{E_i(\Phi)\}_{i = 1}^M$ for fixed semantics experts $\{E_i\}_{i = 1}^M$, then the optimal probability $D^{\ast}$ for the inner maximization in Eq.~\eqref{minmax} is
\begin{equation}
D^{\ast}_i(\Phi^{\prime}, S^{\prime}) = \frac{P^S_i(\Phi^{\prime}, S^{\prime})}{\sum^M_{j = 1}P^S_j(\Phi^{\prime}, S^{\prime})}. \label{prop}
\end{equation}%

\label{proposition}
\end{proposition}
\noindent The proof can be found in Sec. A of the appendix. Building upon Proposition~\ref{proposition}, the ensuing theorem establishes the equivalence between the minimax game in Eq.~\eqref{minmax} and the matching of semantic-based distributions across source domains in Eq.~\eqref{matching}. The detailed proof is provided 
in Sec. B of the appendix.

\begin{theorem}
For a given representation function $f$, if $U$ is the maximum of $-H(D(\Phi, S))$, \emph{i.e.,} $U = \sum_{i = 1}^{M}\sum_{x, y}{P^S_i(x, y)}\log \frac{P^S_i(\Phi, S)}{\sum^M_{j = 1}P^S_j(\Phi, S)}$. Then the solution of Eq.~\eqref{minmax} can be achieved if and only if $P_1^S(\Phi, S) = P_2^S(\Phi, S) = \cdots = P_M^S(\Phi, S)$.
\label{theory}
\end{theorem}

Hence, the objective for matching the semantic-aware distributions across source domains can be formulated as:
\begin{equation}
\begin{split}
    \min_{f} &\max_{D} \ \mathcal{L}_{MDA}, \\
    \text{with} \quad \mathcal{L}_{MDA}&= \mathbb{E}_{P^S(X, Y)}\log D(\Phi, S). \label{adv_loss}
\end{split}
\end{equation}%

\subsection{Optimization Objective}
Together with the classification loss $\mathcal{L}_{cla}$, the overall objective function for our proposed DMDA is:
\begin{equation}
\begin{split}
    &\min_{g, g_a, \{E_i\}_{i = 1}^M, f} \max_{D}\ \mathcal{L},  \\
  \text{with} \quad \mathcal{L} = \mathcal{L}_{cla}& + \alpha \cdot (\mathcal{L}_{MDA} + \mathcal{L}_{exp}) + \beta \cdot \mathcal{L}_{aux}, \label{all_loss}
\end{split}
\end{equation}%
where $\alpha$ and $\beta$ represent the trade-off hyper-parameters. $\mathcal{L}_{cla}$, $\mathcal{L}_{MDA}$, $\mathcal{L}_{exp}$, and $\mathcal{L}_{aux}$ denote the classification loss of the hypothesis model, semantic-aware invariance loss, semantics experts' accuracy loss, and the classification loss of the auxiliary classifier, respectively.

\noindent\textbf{Micro-level Distributions.} In practice, instead of simply concatenating features $\Phi$ and semantics $S$, we employ an element-wise summation of the acquired features and the semantics to construct a surrogate distribution. The idea of element-wise summation is inspired by the operation of position embeddings in Transformer~\cite{vaswani2017attention}, which has been demonstrated to be simple yet effective in embedding the information.

The pseudo-code is provided in Algorithm~\ref{algorithm}.

\begin{table}
\caption{Generalization performance with recognition accuracy (\%) on PACS~\cite{li2017deeper} using ImageNet pre-trained ResNet-50. Higher is better, bold indicates the best performance. \label{PACS}}
    \centering
    \setlength{\tabcolsep}{3pt}
    {
    \begin{tabular}{l|cccc|c}
    \toprule
    Method & Art & Cartoon & Photo & Sketch & Avg.($\uparrow$) \\
    \midrule
    IRM \cite{arjovsky2020invariant}  & $84.8$  & $76.4$ & $96.7$  & $76.1$ & $83.5$ \\
    GroupDRO \cite{sagawa2019distributionally} & $83.5$  & $79.1$ & 96.7  & $78.3$ & $84.4$   \\
    Mixup \cite{yan2020improve}   & $86.1$  & 78.9 & $97.6$  & $75.8$ & $84.6$  \\
    MMD \cite{li2018domain_2}  & $86.1$  & $79.4$ & $96.6$  & $76.5$ & $84.7$ \\
    VREx \cite{krueger2021out}& $86.0$  & $79.1$ & $96.9$  & 77.7 & $84.9$ \\
    RSC \cite{huang2020self}& $85.4$  & $79.7$ & $97.6$  & $78.2$ & $85.2$ \\
    DANN \cite{ganin2016domainadversarial}  & $86.4$  & $77.4$ & 97.3  & $73.5$ & $83.7$ \\
    CDANN \cite{li2018deep}  & $85.0$  & 78.9 & \bfseries98.1  & $76.4$ & $84.6$ \\
    MTL \cite{blanchard2021domain} & $87.5$  & $77.1$ & $96.4$  & $77.3$ & $84.6$  \\
    SagNet \cite{nam2021reducing}& 87.4  & $80.7$ & $97.1$  & 80.0& $86.3$ \\
ARM \cite{zhang2021adaptive}& $86.8$  & $76.8$ & $97.4$  & $79.3$ & $85.1$ \\
    SelfReg$^{\dag}$ \cite{kim2021selfreg}& \bfseries87.9  & $79.4$ & $96.8$  & $78.3$ & $85.6$ \\
    PCL$^{\dag}$ \cite{yao2022pcl}& $87.3$  & $77.5$ & $96.3$  & $83.2$ & $86.1$ \\
   AdaNPC \cite{zhang2023adanpc}& 87.1  & 82.2 & 97.5  & 81.5 & 87.1 \\
    
    FSR \cite{wang2022feature} & 84.4   &  78.3          & 96.0  &  78.1 & 84.2 \\
    IPCL \cite{chen2023instance} & 85.8   &  83.8          & 96.6  &  82.1 & 87.1 \\
    \hline
    DMDA (ours)                                      & 87.1  & \bfseries85.2 & 96.7  & \bfseries83.5 & \bfseries88.1 \\

    \bottomrule
    \end{tabular}}
    \vspace{-1mm}

\end{table}

\section{Experiments}
\label{EXP}

In order to verify the efficiency of the proposed DMDA, we compare it with state-of-the-art approaches in DG, including Invariant Risk Minimization (IRM)~\cite{arjovsky2020invariant}, Group Distributionally Robust Optimization (GroupDRO)~\cite{sagawa2019distributionally}, Interdomain Mixup (Mixup)~\cite{yan2020improve}, Maximum Mean Discrepancy (MMD)~\cite{li2018domain_2}, Variance Risk Extrapolation (VREx)~\cite{krueger2021out}, Representation Self-Challenging (RSC)~\cite{huang2020self}, Domain Adversarial Neural Network (DANN)~\cite{ganin2016domainadversarial}, Conditional Domain Adversarial Neural Network (CDANN)~\cite{li2018deep}, Marginal Transfer Learning (MTL)~\cite{blanchard2021domain}, Style Agnostic Networks (SagNet)~\cite{nam2021reducing}, Adaptive Risk Minimization (ARM)~\cite{zhang2021adaptive}, 
Self-supervised contrastive Regularization (SelfReg)~\cite{kim2021selfreg}, Proxy-based Contrastive Learning (PCL)~\cite{yao2022pcl}, Non-Parametric Classifier
for test-time Adaptation (AdaNPC)~\cite{zhang2023adanpc}, Feature-based Style Randomization~(FSR)~\cite{wang2022feature}, and Instance Paradigm Contrastive Learning~(IPCL)~\cite{chen2023instance}. The performances of those models are from the original literature or DomainBed~\cite{gulrajani2020search_2}.
While $^{\dag}$ in the tables means the reproduced results without the ensembling technique for a fair comparison.

\subsection{Datasets}
Following previous DG protocols~\cite{li2023sparse, kim2021selfreg, cha2021swad, gulrajani2020search_2}, we validate our proposed methods on five widely-used datasets: (1) \textbf{PACS}~\cite{li2017deeper} comprises four distinct domains: Photo, Art, Cartoon, and Sketch, with each domain encompassing seven classes. (2) \textbf{VLCS}~\cite{fang2013unbiased} contains four standard datasets: Caltech~\cite{fei2004learning}, LabelMe~\cite{russell2008labelme}, SUN~\cite{choi2010exploiting}, and VOC~\cite{everingham2010pascal}. In each dataset, there are five categories: bird, car, chair, dog, and person. (3) \textbf{OfficeHome}~\cite{venkateswara2017deep} consists of four diverse domains: Art, Clipart, Product, and Real-World, each containing 65 classes. (4) \textbf{TerraIncognita}~\cite{beery2018recognition} showcases photographs of wild animals captured at various locations: L100, L38, L43, and L46, each of which includes 10 classes. (5) \textbf{DomainNet}~\cite{peng2019moment}constitutes a substantial dataset, encompassing 586,575 images across six domains, with each domain comprising 345 classes.

\begin{table}
\caption{Generalization performance with recognition accuracy (\%) on VLCS~\cite{fang2013unbiased} using ImageNet pre-trained ResNet-50. Higher is better, bold indicates the best performance.    \label{VLCS}}
    \centering
    \setlength{\tabcolsep}{3pt}
    {
    \begin{tabular}{l|cccc|c}
    \toprule
    Method & Caltech & LabelMe & SUN & VOC & Avg.($\uparrow$) \\
    \midrule
    IRM \cite{arjovsky2020invariant}  & $98.6$  & 64.9 & $73.4$  & $77.3$ & $78.5$ \\
    GroupDRO \cite{sagawa2019distributionally} & $97.3$  & $63.4$ & 69.5  & $76.7$ & $76.7$   \\
    Mixup \cite{yan2020improve}  & 98.3  & 64.8 & $72.1$  & $74.3$ & $77.4$ \\
    MMD \cite{li2018domain_2}  & $97.7$  & $64.0$ & $72.8$  & 75.3 & $77.5$  \\
    VREx \cite{krueger2021out} & $98.4$  & $64.4$ & 74.1  & 76.2 & $78.3$\\
    RSC \cite{huang2020self} & $97.9$  & $62.5$ & $72.3$  & $75.6$ & $77.1$\\
    DANN \cite{ganin2016domainadversarial}  & \bfseries99.0  & $65.1$ &73.1  & $77.2$ & $78.6$ \\
    
    CDANN \cite{li2018deep} & $97.1$  & 65.1 & $70.7$  & 77.1 & 77.5\\
    MTL \cite{blanchard2021domain} & $97.8$  & $64.3$ & $71.5$  & $75.3$ & $77.2$ \\
    SagNet \cite{nam2021reducing} & 97.9  & $64.5$ & $71.4$  & 77.5 & $77.8$\\
    ARM \cite{zhang2021adaptive}& 98.7  & $63.6$ & $71.3$  & $76.7$ & $77.6$\\
    SelfReg$^{\dag}$ \cite{kim2021selfreg}& $96.7$  & \bfseries65.2 & $73.1$  & $76.2$ & $77.8$ \\
    PCL$^{\dag}$ \cite{yao2022pcl}& $96.6$  & $58.1$ & $72.4$  & $75.2$ & $75.6$ \\
    AdaNPC \cite{zhang2023adanpc}& 98.9  & 64.5 & 73.5  & 75.6 & 78.1 \\
   FSR \cite{wang2022feature} & 89.6   &  64.1          & 70.8  &  75.0 & 74.9 \\
   IPCL \cite{chen2023instance} & 97.7   &  63.8          & 73.8  &  78.1 & 78.4 \\
    \hline
    DMDA (ours) & 98.2  & $64.3$ & \bfseries74.4  & \bfseries80.8 & \bfseries79.4 \\

    \bottomrule
    \end{tabular}}
   \vspace{-1mm} 

\end{table}

\begin{table}
\caption{Generalization performance with recognition accuracy (\%) on OfficeHome~\cite{venkateswara2017deep} using ImageNet pre-trained ResNet-50. Higher is better, bold indicates the best performance.    \label{office_home}}
    \centering
    \setlength{\tabcolsep}{3pt}
    {
    \begin{tabular}{l|cccc|c}
    \toprule
    Method & Art & Clipart & Product & Real-World & Avg.($\uparrow$) \\
    \midrule
    IRM \cite{arjovsky2020invariant}  & $58.9$  & $52.2$ & $72.1$  & $74.0$ & $64.3$\\
    GroupDRO \cite{sagawa2019distributionally}& $60.4$  & $52.7$ & 75.0  & $76.0$ & $66.0$  \\
    Mixup \cite{yan2020improve} & $62.4$  & 54.8 & 76.9  & 78.3 & $68.1$ \\
    MMD \cite{li2018domain_2}   & $60.4$  & 53.3& $74.3$  & $77.4$ & $66.4$\\
    VREx \cite{krueger2021out}  & $60.7$  & $53.0$ & $75.3$  & 76.7 & $66.4$\\
    RSC \cite{huang2020self}    & $60.7$  & $51.4$ & $74.8$  & $75.1$ & $65.5$ \\
    DANN \cite{ganin2016domainadversarial}  & $59.9$  & $53.0$ & 73.6  & $76.9$ & $65.9$ \\
    
    CDANN \cite{li2018deep}     & $61.5$  & 50.4 & $74.4$  & $76.6$ & 65.7 \\
    MTL \cite{blanchard2021domain}  & $61.5$  & $52.4$ & $74.9$  & $76.8$ & $66.4$  \\
    SagNet \cite{nam2021reducing}      & 63.4  & 54.8 & $75.8$  & 78.3 & $68.1$\\
    ARM \cite{zhang2021adaptive}  & $58.9$  & $51.0$ & $74.1$  & $75.2$ & $64.8$ \\
    SelfReg$^{\dag}$ \cite{kim2021selfreg}& $63.6$  & 53.1 & $76.9$  & $78.1$ & $67.9$ \\
    PCL$^{\dag}$ \cite{yao2022pcl}& $62.7$  & $54.0$ & $76.9$  & \bfseries78.5 & $68.0$ \\
   AdaNPC \cite{zhang2023adanpc}& 62.9  & 52.3 & 75.1  & 75.6 & 66.5 \\
   FSR \cite{wang2022feature}  & 59.7   &  55.9          & 74.7  &  74.9 & 66.3 \\
   IPCL \cite{chen2023instance} & 65.1   &  \bfseries56.8          & 77.3 &  77.1 & 69.1 \\
    \hline
    DMDA (ours)  & \bfseries66.1  & 56.1 & \bfseries77.3  & 77.9 & \bfseries69.4\\

    \bottomrule
    \end{tabular}}

\end{table}
\subsection{Implementation Details}
Adhering to established DG protocols~\cite{cha2021swad, gulrajani2020search_2}, we employ ImageNet~\cite{ deng2009imagenet} pre-trained ResNet-50~\cite{he2016deep} as the backbone, with the last FC layer serving as the label predictor and the preceding layers functioning as the feature extractor. We utilize three FC layers to construct the distribution approximator which receives inputs through a gradient reverse layer~(GRL)~\cite{ganin2016domainadversarial}, and the output dimensions are 256, 256, and the number of source domains, respectively. Our model undergoes training for 15k iterations for DomainNet and 5k iterations for others, utilizing SGD with a momentum of 0.9, weight decay set to 5e-4. The learning rate is decayed by a factor of 0.1 at 70\% and 90\% of the total epochs. The batch size is fixed at 32 for each domain. The adaptation parameter in GRL is initialized at 0 and gradually transitions to 1 following~\cite{ganin2016domainadversarial}. For additional hyperparameter details, please refer to TABLE~\ref{para}.
\begin{table}[]
    \centering
    \caption{Hyperparameter search space, with $m = q /100$.}
    \begin{tabular}{c|c}
    \toprule
         Hyperparameter &  Search space\\
         \midrule
         $\alpha$ &  [1] \\
         $\beta$ &  [3, 5] \\
         $m$ &  [0.5, 0.4, 0.3] \\
         learning rate & [1e-3, 5e-4]\\
         \bottomrule
    \end{tabular}
    \label{para}
\end{table}

To ensure a fair comparison, we adhere to the training-domain validation evaluation protocol following DomainBed~\cite{gulrajani2020search_2}. This entails cyclically selecting one domain as the target domain, with the remaining domains serving as source domains. Within each source domain, we execute an 80\%/20\% train/validation partition, whereby the validation portions across all source domains collectively compose the validation dataset used for model validation and selection.

\subsection{Comparison Results with State-of-the-Art Techniques}

The results on PACS are presented in TABLE~\ref{PACS}.
While our approach may not consistently achieve the highest accuracy in specific scenarios, it demonstrates significant improvements across most scenarios. Moreover, the highest average object recognition accuracy which outperforms the SOTA method IPCL by 1.1\% attests to the efficacy of DMDA for DG. Notably, DMDA excels in enhancing the generalization performance in hard-to-transfer domains where style markedly diverges from the other domains. It exhibits a notable superiority of 1.7\% on `Cartoon' and 1.3\% on `Sketch' compared to IPCL. This observation underscores DMDA's capacity to cultivate superior representations for effective generalization to novel environments.

TABLE~\ref{VLCS} reports the outcomes for VLCS.
DMDA achieves the best average generalization performance and demonstrates substantial improvements across most scenarios when compared to alternative methods. Additionally, in hard-to-transfer domains, DMDA outperforms the SOTA method DANN by 4.7\% on `VOC' and 1.8\% on `SUN'. This outcome underscores the superior efficacy of DMDA, even in the face of the heightened generalization challenges posed by scene-centric images within VLCS.

\begin{table}
\caption{Generalization performance on TerraIncognita~\cite{beery2018recognition} using ImageNet pre-trained ResNet-50. Higher is better, bold indicates the best performance out of all compared methods.    \label{terra}}
    \centering
    \setlength{\tabcolsep}{5pt}    {
    \begin{tabular}{l|cccc|c}
    \toprule
    Method & L100 & L38 & L43 & L46 & Avg.($\uparrow$) \\
    \midrule
    IRM \cite{arjovsky2020invariant}  & $54.6$  & $39.8$ & $56.2$  & $39.6$ & $47.6$\\
    GroupDRO \cite{sagawa2019distributionally}& $41.2$  & $38.6$ & 56.7  & $36.4$ & $43.2$  \\
    Mixup \cite{yan2020improve} & \bfseries59.6  & 42.2 & 55.9  & 33.9 & $47.9$ \\
   MMD \cite{li2018domain_2}   & $41.9$  & 34.8& $57.0$  & $35.2$ & $42.2$\\
    VREx \cite{krueger2021out}  & $48.2$  & $41.7$ & $56.8$  & 38.7 & $46.4$\\
    RSC \cite{huang2020self}    & $50.2$  & $39.2$ & $56.3$  & $40.8$ & $46.6$ \\
    DANN \cite{ganin2016domainadversarial}  & $51.1$  & $40.6$ & 57.4  & $37.7$ & $46.7$ \\
    
    CDANN \cite{li2018deep}    & $47.0$  & 41.3 & $54.9$  & $39.8$ & $45.8$ \\
   MTL \cite{blanchard2021domain}  & $49.3$  & $39.6$ & $55.6$  & $37.8$ & $45.6$  \\
    SagNet \cite{nam2021reducing}      & 53.0  & 43.0 & \bfseries57.9  & 40.4 & 48.6\\
    ARM \cite{zhang2021adaptive}  & $49.3$  & $38.3$ & $55.8$  & $38.7$ & $45.5$ \\
    SelfReg$^{\dag}$ \cite{kim2021selfreg}& 48.8  & 41.3 & $57.3$  & $40.6$ & $47.0$ \\
    PCL$^{\dag}$ \cite{yao2022pcl}& $41.6$  & $42.8$ & $52.9$  & $40.1$ & $44.3$ \\
   AdaNPC \cite{zhang2023adanpc}& 56.7  & \bfseries48.5 & 47.5  & 35.4 & 47.0 \\
     FSR \cite{wang2022feature}  & 57.4   &  37.8          & 54.8  &  36.1 & 46.5 \\
     IPCL \cite{chen2023instance} & 50.9 & 41.3   &  54.6          & 40.6 & 46.8 \\
    \hline
    DMDA (ours)  & 55.9  & 41.7 & $56.8$  & \bfseries43.6 & \bfseries49.5\\

    \bottomrule
    \end{tabular}}

\end{table}

\begin{table}
\caption{Generalization performance on DomainNet~\cite{peng2019moment} using ImageNet pre-trained ResNet-50. Higher is better, bold indicates the best performance out of all compared methods.    \label{domain_net}}
    \centering
    \setlength{\tabcolsep}{1pt}
    {
    \begin{tabular}{l|cccccc|c}
    \toprule
    Method & Clipart & Infograph & Painting & Quickdraw & Real & Sketch & Avg.($\uparrow$) \\
    \midrule
    IRM [70]  & 48.5  & 15.0 & 38.3  & 10.9 & 48.2 & 42.3 & 33.9\\
    GroupDRO [71]& 47.2  & 17.5 & 33.8  & 9.3 & 51.6 & 40.1 & 33.3  \\
    Mixup [72] & 55.7  & 18.5 & 44.3  & 12.5 & 55.8 & 48.2 & 39.2 \\
    MMD [44]   & 32.1  & 11.0 & 26.8  & 8.7 & 32.7 & 28.9 & 23.4\\
    VREx [73]  & 47.3  & 16.0 & 35.8  & 10.9 & 49.6 & 42.0 & 33.6\\
    RSC [74]    & 55.0  & 18.3 & 44.4  & 12.2 & 55.7 & 47.8 & 38.9 \\
    DANN [30]  & 53.1  & 18.3 & 44.2  & 11.8 & 55.5 & 46.8 & 38.3 \\
    
    CDANN [52]     & 54.6  & 17.3 & 43.7  & 12.1 & 56.2 & 45.9 & 38.3 \\
    MTL [75]  & 57.9  & 18.5 & 46.0  & 12.5 & 59.5 & 49.2 & 40.6  \\
    SagNet [76]      & 57.7  & 19.0 & 45.3  & 12.7 & 58.1 & 48.8 & 40.3\\
    ARM [77]  & 49.7 & 16.3 & 40.9  & 9.4 & 53.4 & 43.5 & 35.5 \\
    SelfReg$^{\dag}$ [57]& 60.7  & 21.6 & 49.4  & 12.7 & 60.7 & 51.7 & 42.8 \\
    PCL$^{\dag}$ [48]& 62.5  & 21.1 & 49.5  & 14.1 & 61.3 & 51.8 & 43.4 \\
   AdaNPC [78]& 59.3  & 22.2 & 48.3  & 14.3 & 61.0 & 51.4 & 42.8 \\
   FSR \cite{wang2022feature}  &  59.0          & 18.2         &46.7          & 15.0 & 55.8 & 49.5         &  40.7\\
  IPCL \cite{chen2023instance} & 55.3   & 18.6          & 39.4 & 11.9      & \bfseries64.7 & 53.9 & 40.6 \\
    \hline
    DMDA (ours)  & \bfseries65.8  & \bfseries22.5 & \bfseries51.9  & \bfseries15.5 & 64.2 & \bfseries54.7 & \bfseries45.8\\

    \bottomrule
    \end{tabular}}

\end{table}

The generalization performances for OfficeHome are detailed in TABLE~\ref{office_home}.
DMDA consistently demonstrates superior performance in the majority of scenarios, along with achieving the highest average object recognition accuracy. A remarkable aspect of DMDA is its substantial performance enhancement in hard-to-transfer domains like `Art' and `Clipart'. This notable improvement underscores the validity of our approach, which strives to cultivate improved representations characterized by superior discriminability and generalizability.

TABLE~\ref{terra} displays the findings of TerraIncognita.
DMDA outperforms state-of-the-art methods in terms of average object recognition accuracy. Furthermore, our model exhibits a significant improvement of 7.9\% compared to the SOTA method in the most challenging scenario, namely, the `L46' domain. This observation highlights the inherent capability of our method to acquire discriminative representations across domains. 

The outcomes on the more challenging benchmark, DomainNet, are documented in TABLE~\ref{domain_net}, demonstrating the superior generalization performance of our proposed DMDA in five out of six scenarios. Furthermore, DMDA surpasses the state-of-the-art method PCL by 5.5\% in terms of average generalization performance. These results substantiate our assertion that the proposed DMDA has the potential to enhance both feature discrimiability and discriminability.

\begin{figure*}[t!]
        \centering
        \subfloat[ERM]{\includegraphics[width=1.6in]{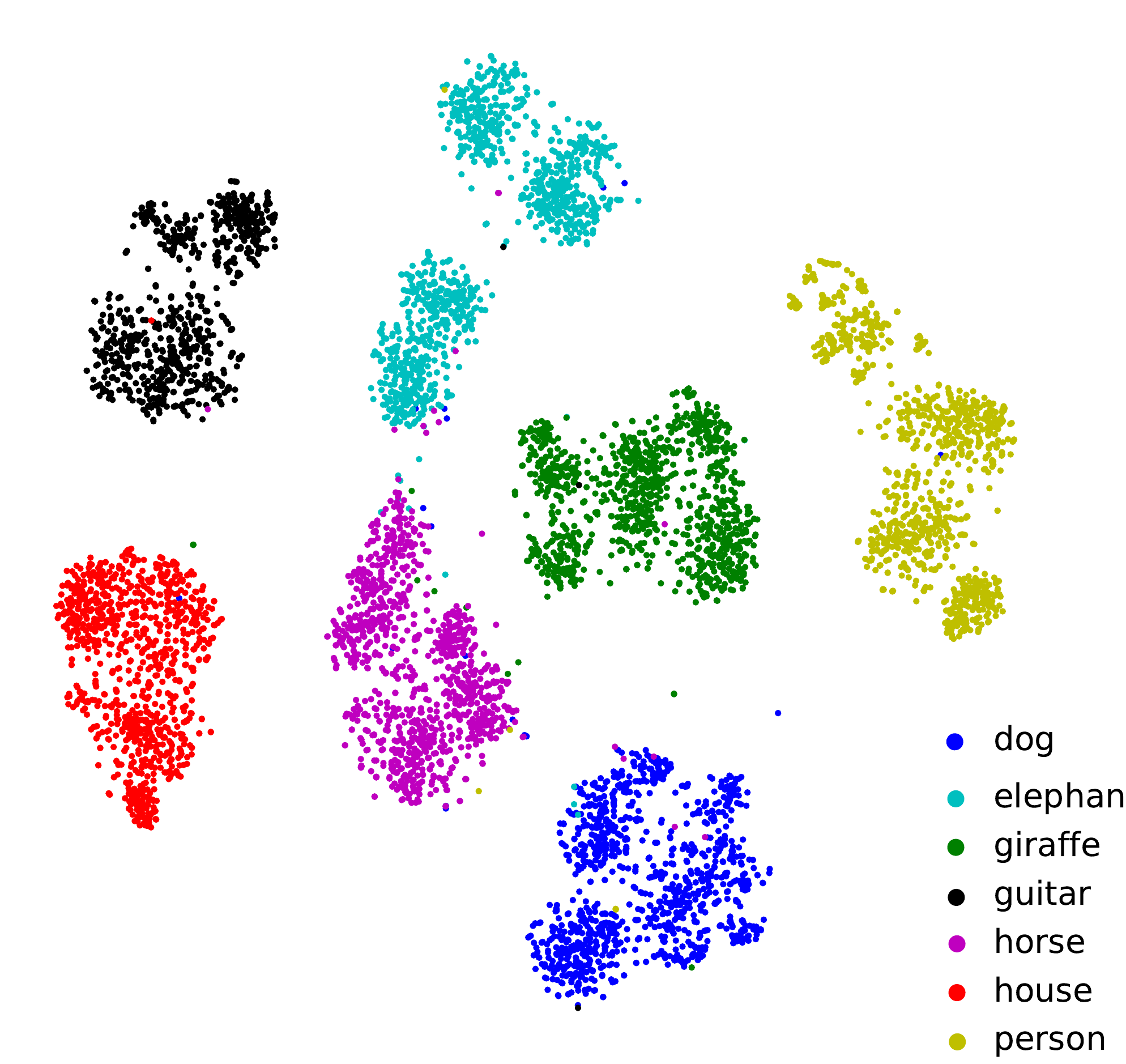}}
         \hspace{0.3cm}
        \subfloat[SCP]{\includegraphics[width=1.6in]{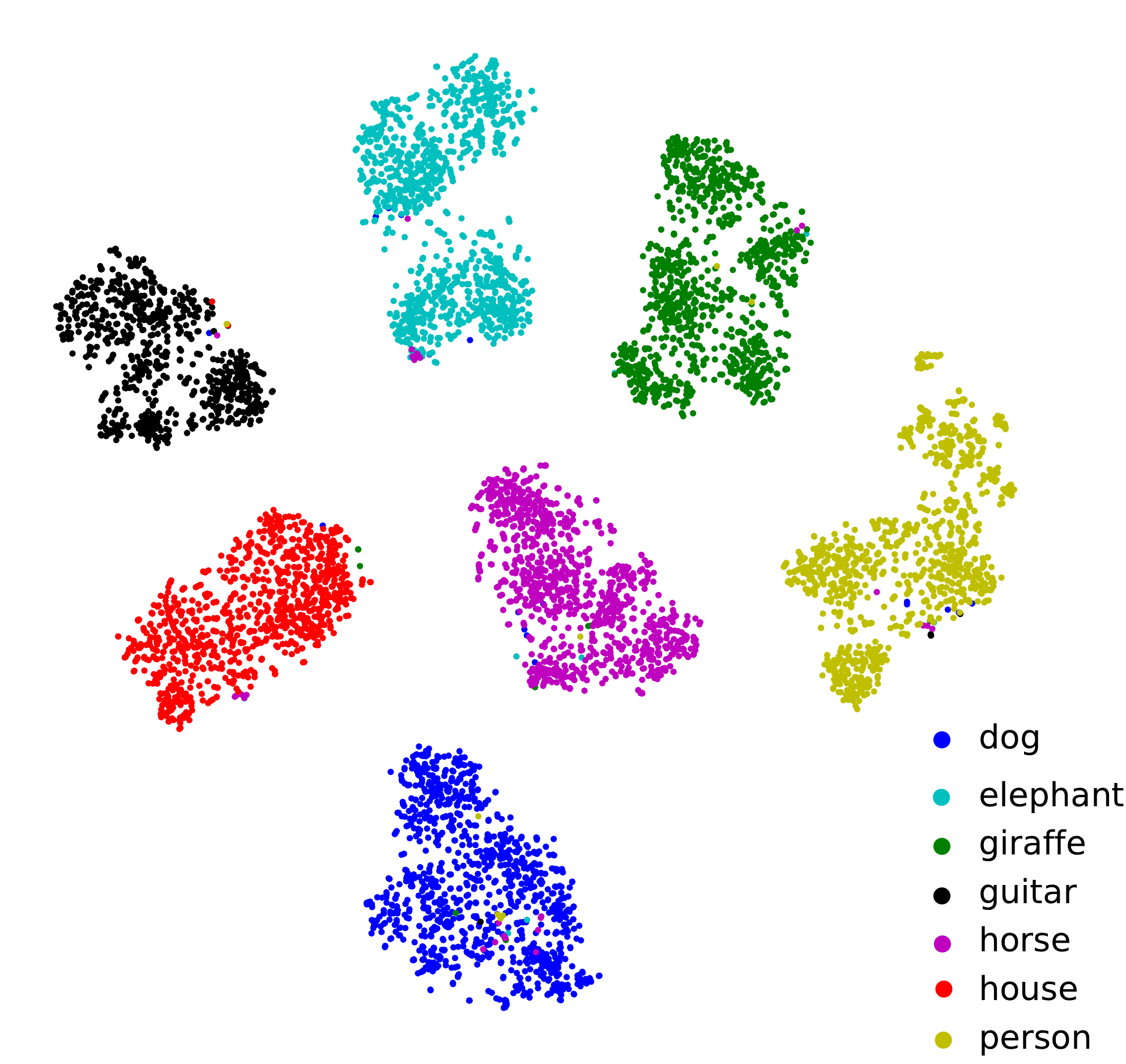}}
         \hspace{0.3cm}
        \subfloat[MDA]{\includegraphics[width=1.6in]{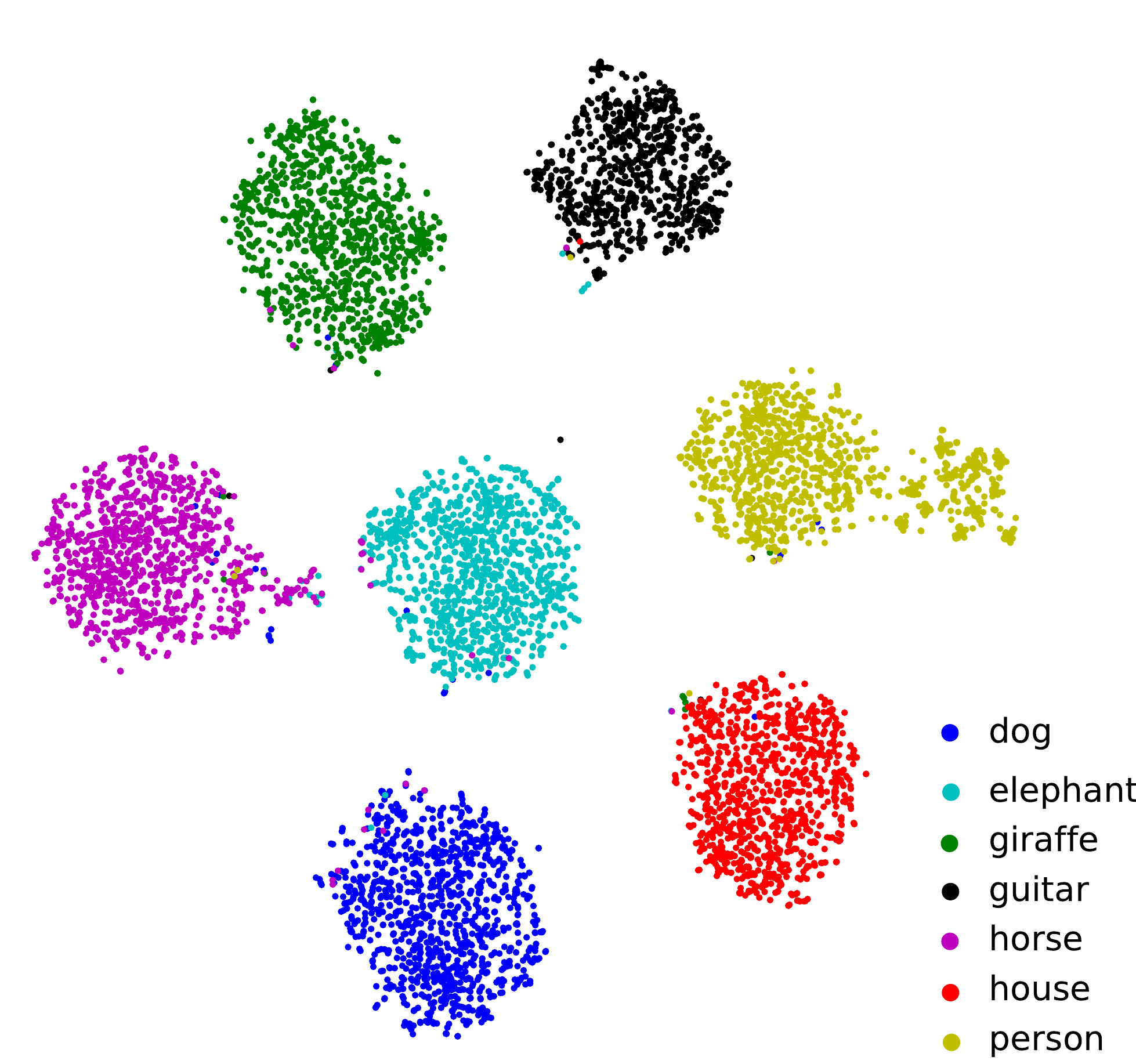}}
         \hspace{0.3cm}
        \subfloat[DMDA]{\includegraphics[width=1.6in]{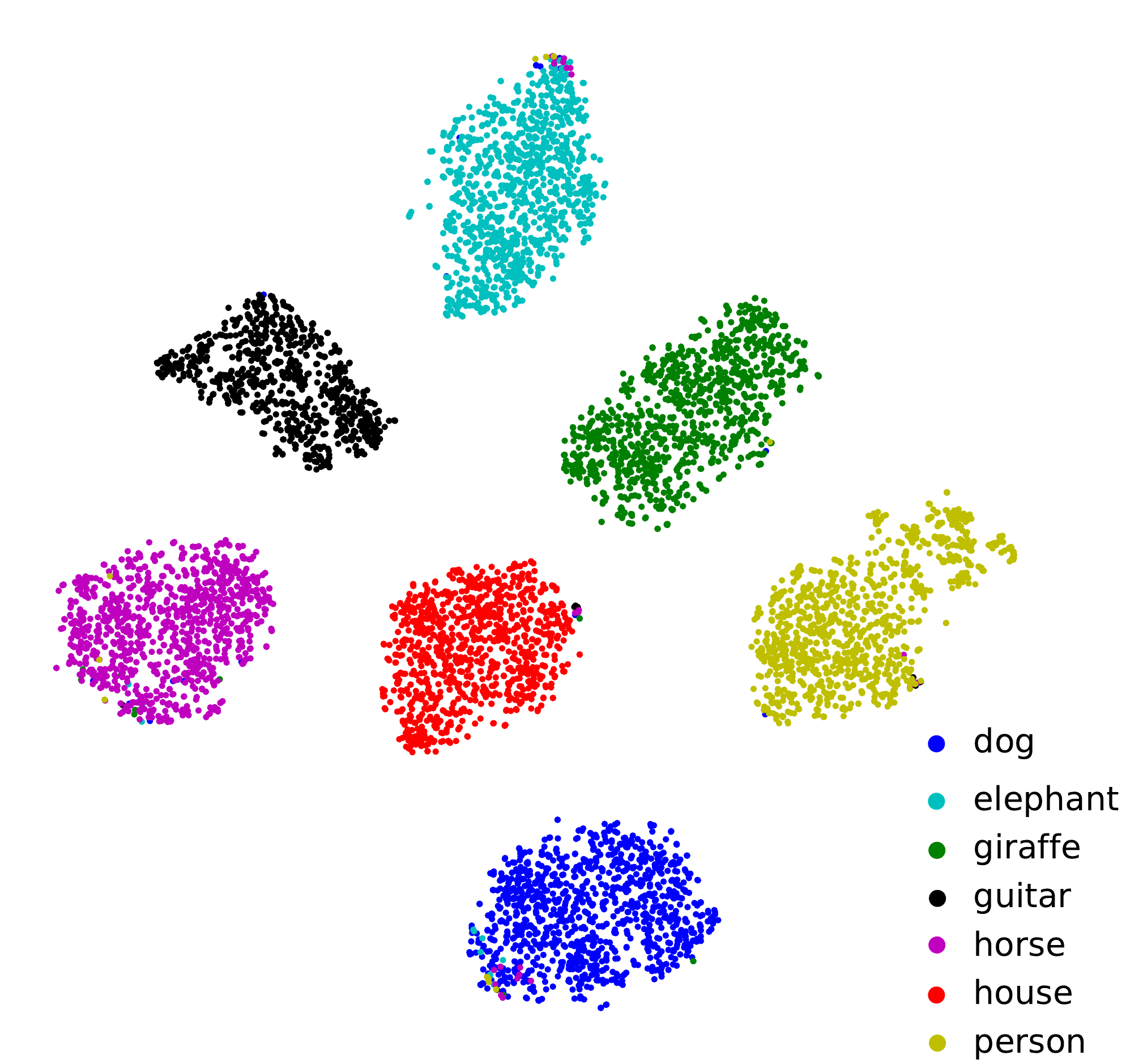}}
        \caption{Visualizations with t-SNE embeddings~\cite{van2008visualizing} depicting distinct classes' representations produced by (a) ERM, (b) SCP, (c) MDA, and (d) the combination DMDA, respectively, with `Photo' as the target domain. Zoom in for details.        \label{compa_TSNE}}
\vspace{-3mm}
    \end{figure*}

\begin{table}
\caption{Ablation study of the our DMDA on diverse benchmarks. Higher is better, 
    bold indicates the best performance.    \label{SDAT}}
    \centering
    \setlength{\tabcolsep}{1pt}
    {
    \begin{tabular}{l|ccccc}
    \toprule
    Method & PACS & VLCS & OfficeHome & TerraIncognita & DomainNet  \\
    \midrule
    ERM      &  85.5            & 77.5           & 66.5            & 46.1    & 43.0       \\
    \ w/ SCP   &  86.7           & 78.4          & 68.7           & 48.5 & 45.1          \\
    \ w/ MDA & 86.9   & 78.6          & 67.7  & 48.9 & 44.2           \\
    \hline
    DMDA (ours)     & \bfseries88.1            & \bfseries79.4 & \bfseries69.4           & \bfseries49.5 & \bfseries45.8  \\
    \bottomrule
    \end{tabular}}

    \end{table}

\begin{table}
\caption{Comparison between Dropout~\cite{srivastava2014dropout} and SCP on diverse benchmarks. Higher is better, 
    bold indicates the best performance.    \label{dropout}}
    \centering
    \setlength{\tabcolsep}{1pt}
    {
    \begin{tabular}{l|ccccc}
    \toprule
    Method & PACS & VLCS & OfficeHome & TerraIncognita & DomainNet  \\
    \midrule
    Dropout~\cite{srivastava2014dropout} & 84.9   & 78.3         & 64.8  & 47.4 & 43.7           \\
    SCP (ours)     & \bfseries88.1            & \bfseries79.4 & \bfseries69.4           & \bfseries49.5 & \bfseries45.8  \\
    \bottomrule
    \end{tabular}}    
    \vspace{-2mm}
    \end{table}

\subsection{Ablation Study}

\noindent\textbf{Effectiveness of Each Component.}
We undertake ablation studies on five benchmarks to assess the individual effectiveness of SCP and MDA. TABLE~\ref{SDAT} presents the generalization performance as we incrementally integrate SCP and MDA into ERM. As can be observed, both SCP and MDA contribute positively to enhanced generalization capabilities. The optimal performance is realized when SCP and MDA are combined, underscoring the indispensability of both SCP and MDA in improving generalization performance.

\noindent
\textbf{The t-SNE Visualization of Features.}
With the collaborative integration of SCP and MDA, the learned features exhibit the superiority of both high discriminability and generalizability. Fig.~\ref{compa_TSNE} depicts the t-SNE embeddings of the learned features with ERM, SCP, MDA, and the cooperation DMDA, respectively. As observed, both SCP and MDA manifest superior representations in comparison to ERM. Moreover, the cooperation exhibits better representations with heightened intra-class compactness in comparison to SCP and superior inter-class separation compared to MDA, demonstrating the necessity of both SCP and MDA.

\subsection{Empirical Analysis}
In this section, we conduct empirical analysis to investigate the impact of different modules in DMDA and compare it with existing distribution matching methods to assess the effectiveness of our method. The analysis aims to shed light on the reasons behind our approach's strong performance.

\begin{figure*}[t!]
        \centering
        \subfloat[Art]{\includegraphics[width=1.6in]{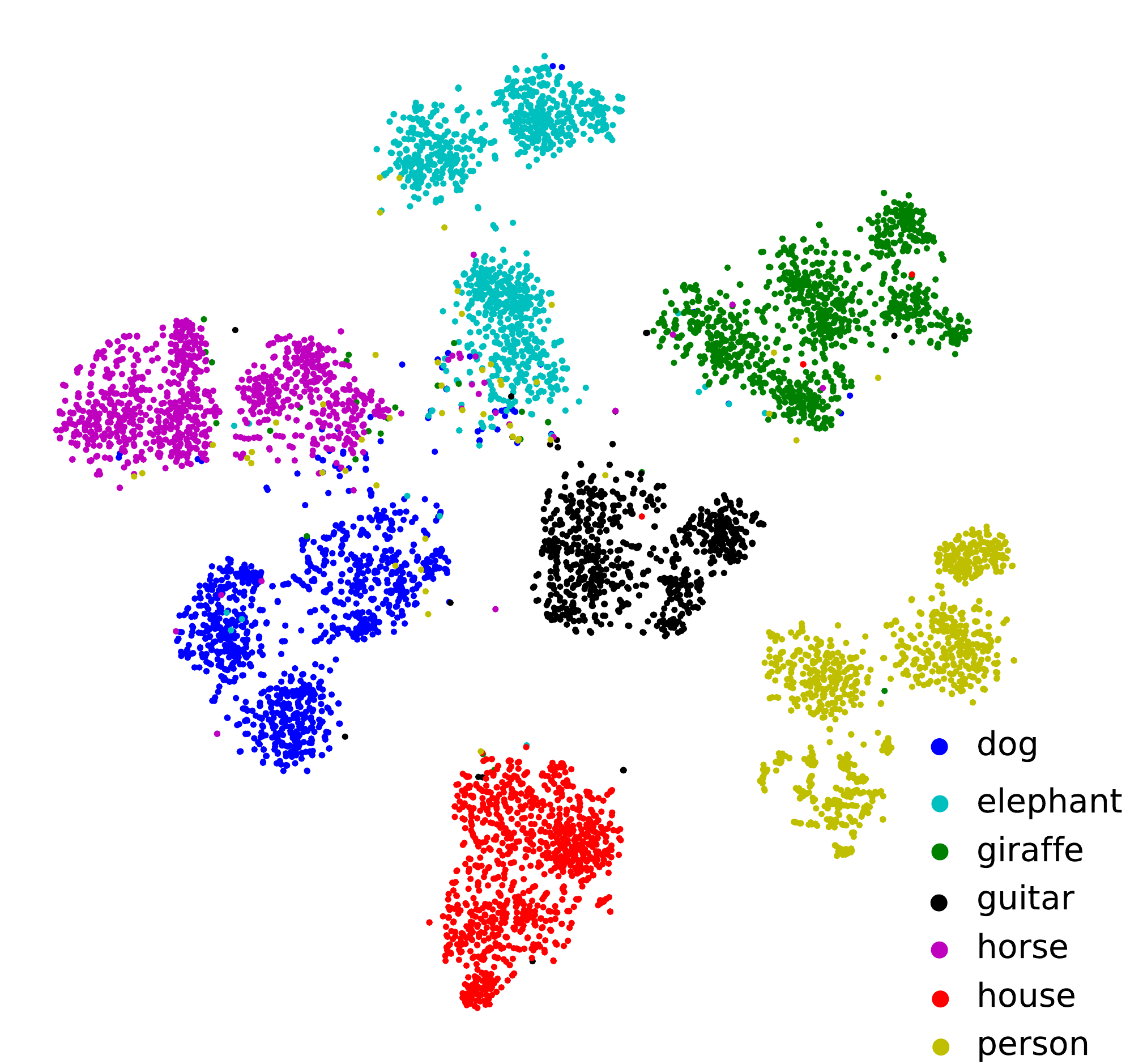}}
         \hspace{0.3cm}
        \subfloat[Cartoon]{\includegraphics[width=1.6in]{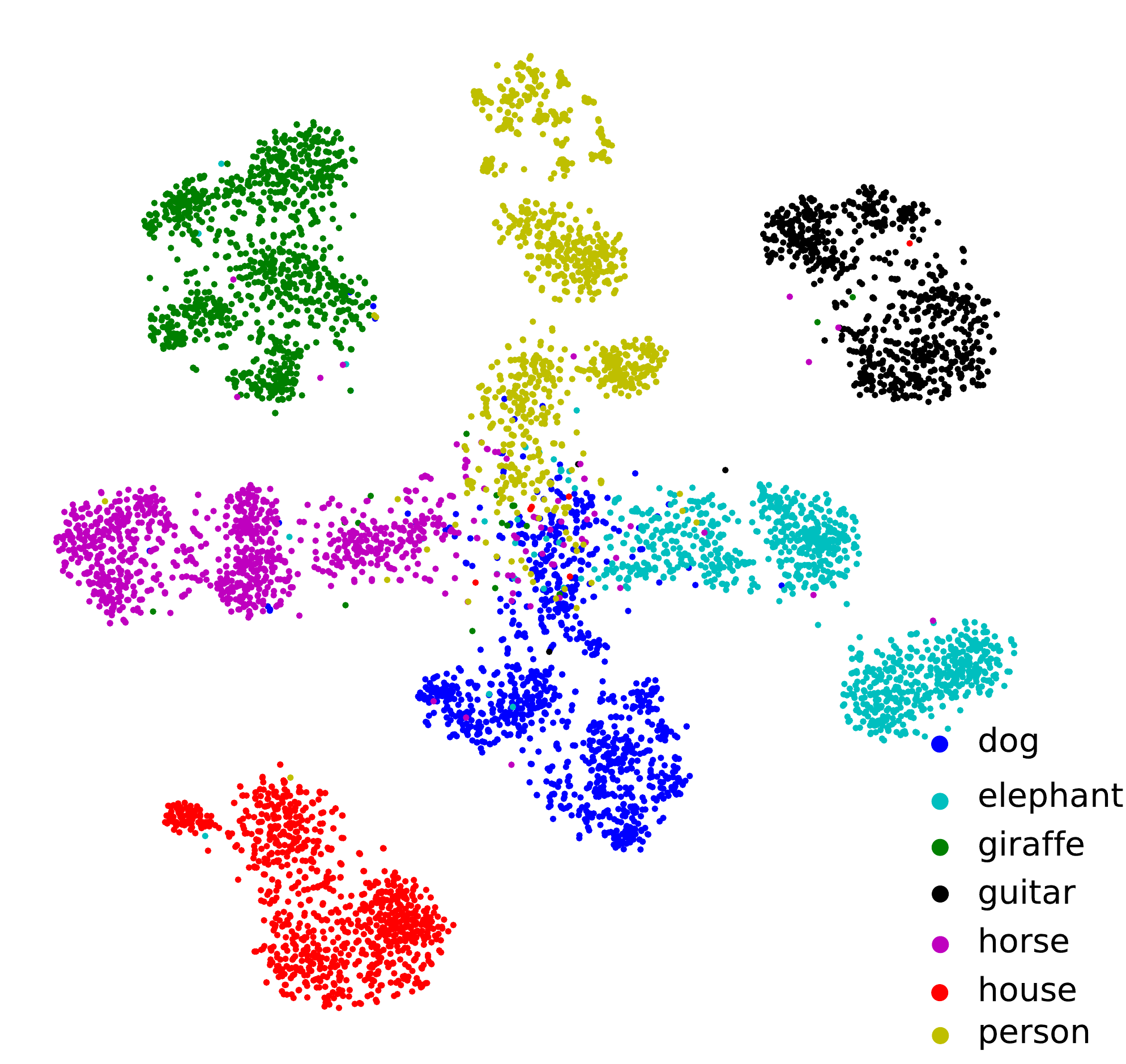}}
         \hspace{0.3cm}
        \subfloat[Photo]{\includegraphics[width=1.6in]{Figure/ERM_MASK/photo.pdf}}
         \hspace{0.3cm}
        \subfloat[Sketch]{\includegraphics[width=1.6in]{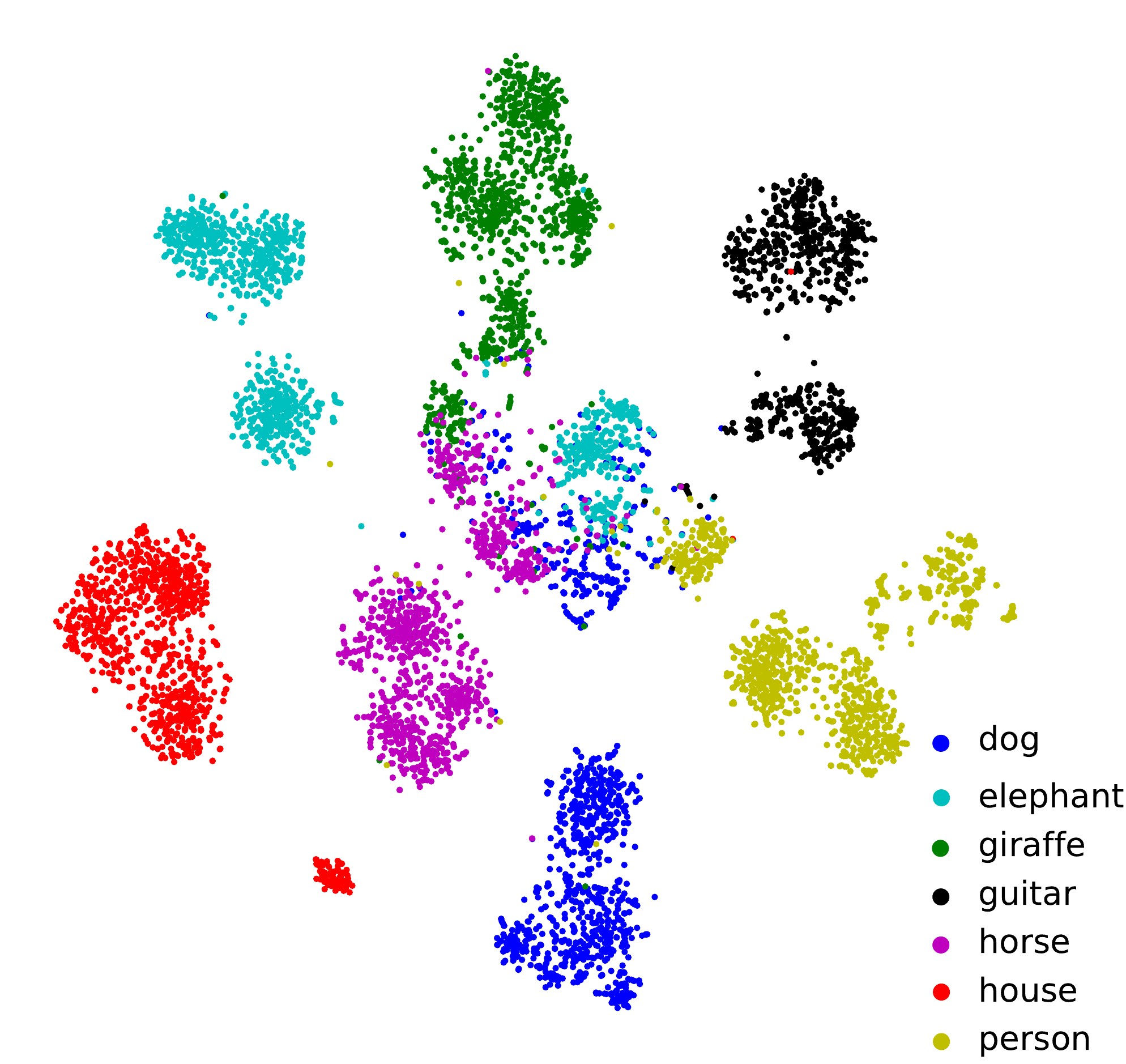}}
        \caption{Visualizations with t-SNE embeddings~\cite{van2008visualizing} depicting representations of distinct classes of ERM, where (a) Art, (b) Cartoon, (c) Photo, and (d) Sketch are individually selected as the target domain.  Zoom in for details.        \label{ERM_TSNE}}

         \vspace{-1mm}
    \end{figure*}
   
\begin{figure*}[t!]
        \centering
        
        \subfloat[Art]{\includegraphics[width=1.6in]{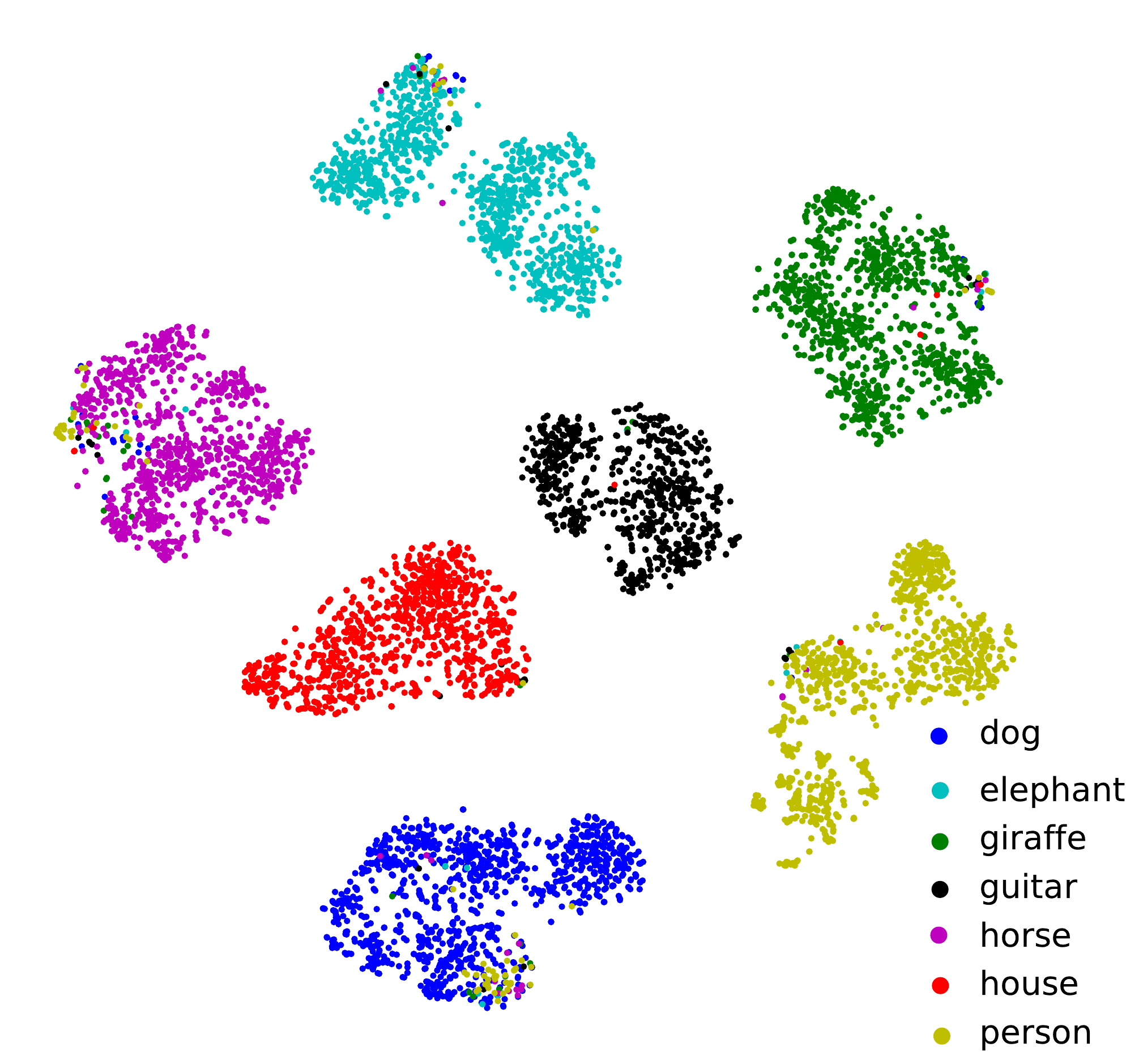}}
         \hspace{0.3cm}
        \subfloat[Cartoon]{\includegraphics[width=1.6in]{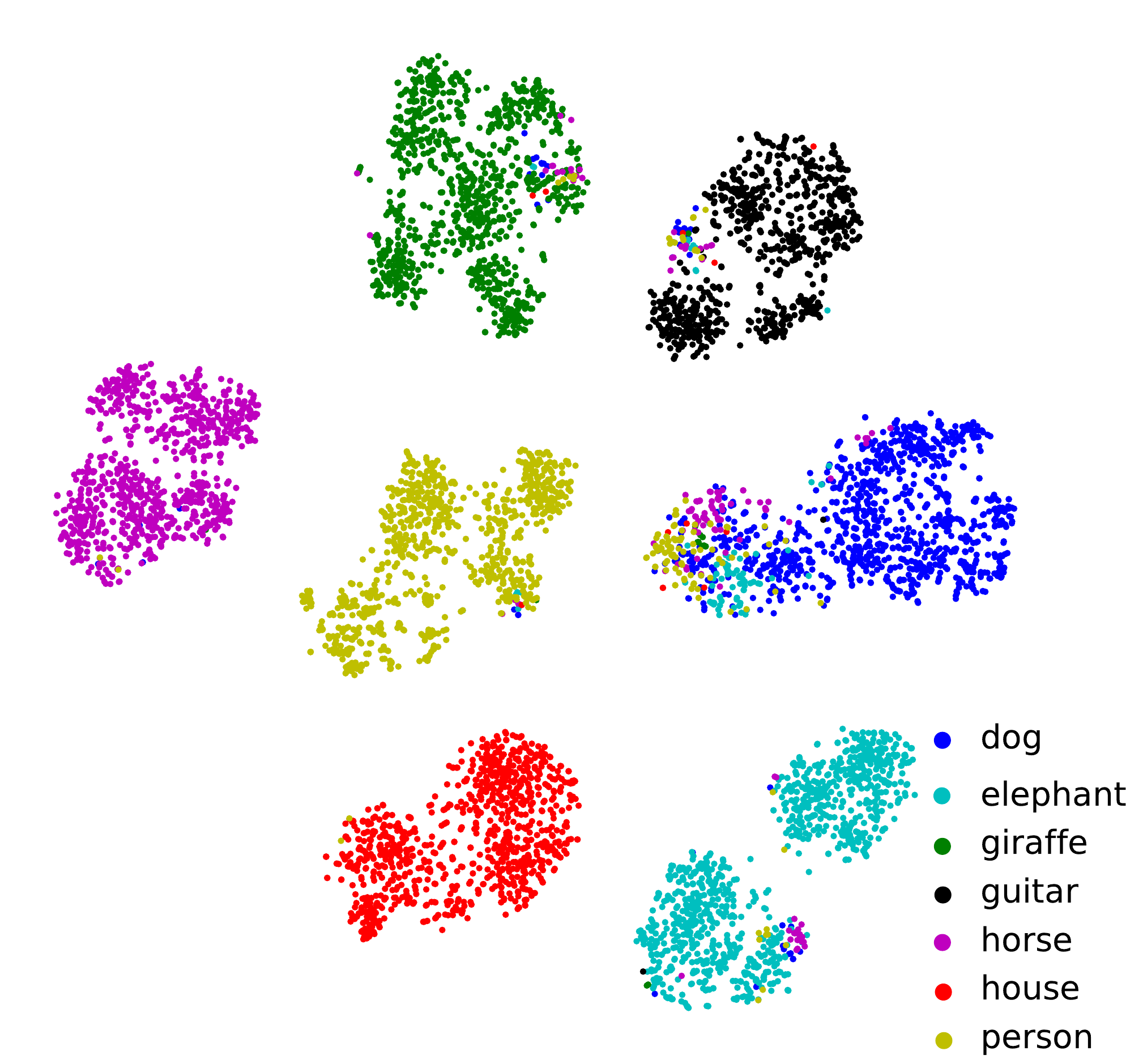}}
         \hspace{0.3cm}
        \subfloat[Photo]{\includegraphics[width=1.6in]{Figure/ERM_MASK/mask_photo.pdf}}
         \hspace{0.3cm}
        \subfloat[Sketch]{\includegraphics[width=1.6in]{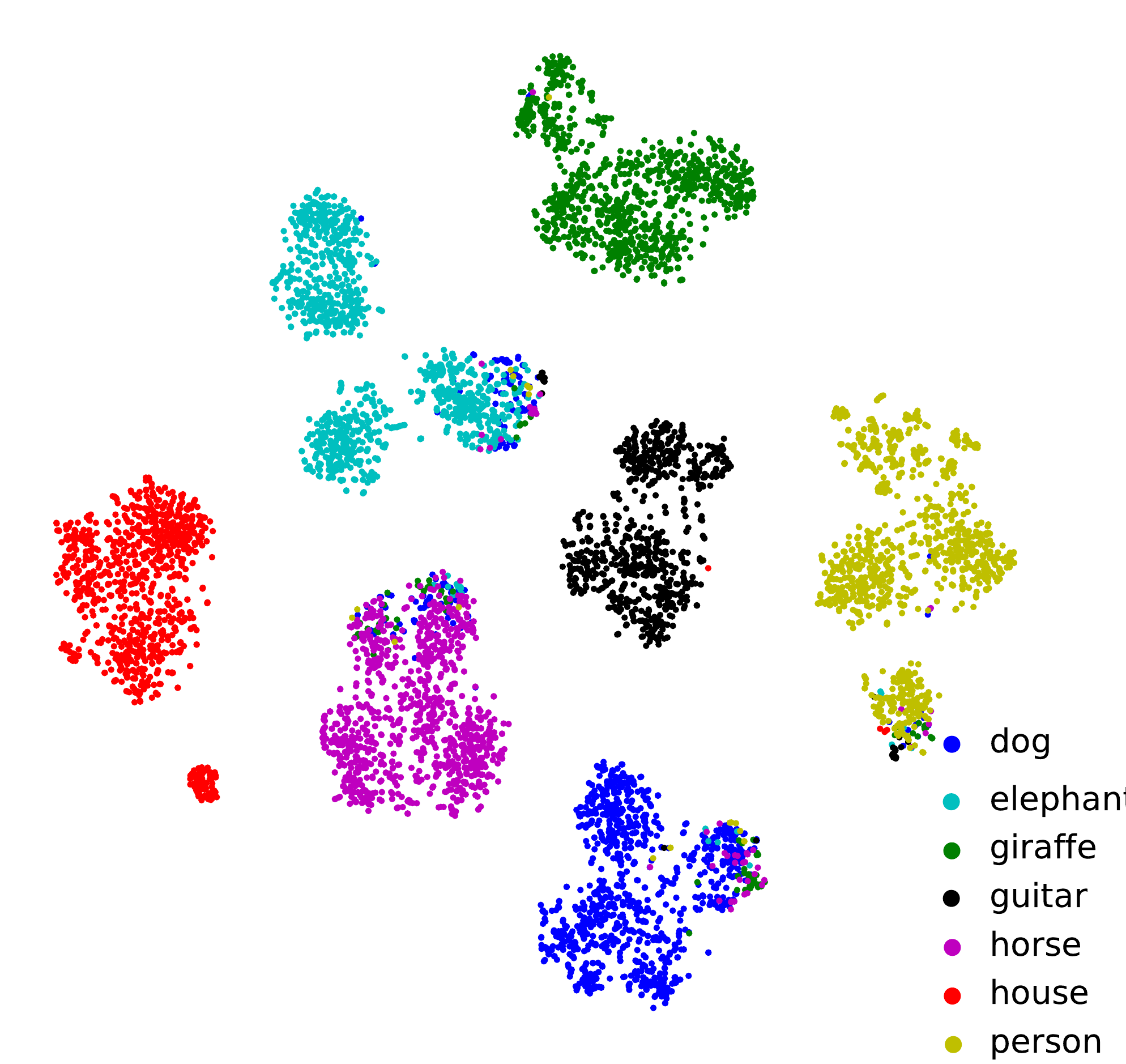}}
        \centering
        \caption{Visualizations using t-SNE embeddings~\cite{van2008visualizing} representing features of SCP-based ERM, with (a) Art, (b) Cartoon, (c) Photo, and (d) Sketch individually chosen as the target domain. The pruned features exhibit an clustering effect. Zoom in for details.       \label{MASK_TSNE}}
 
        \vspace{-1mm}
    \end{figure*}

\begin{table}
\caption{Integration of SCP with distribution matching methods on PACS. Higher is better, 
    bold indicates the best performance.}
    \centering
    \setlength{\tabcolsep}{3pt}
    {
    \begin{tabular}{l|cccc|c}
    \toprule
    Method & Art & Cartoon & Photo & Sketch & Avg.($\uparrow$) \\
    \midrule
    DANN \cite{ganin2016domainadversarial}     &  86.4            & 77.4           & \bfseries97.3            & 73.5           & 83.7 \\
    \ w/ SCP   &  \bfseries87.0           & \bfseries83.0          & 96.9           & \bfseries76.0          &  \bfseries85.7\\
    \hline
    CDANN \cite{li2018deep}     &  85.0            & 78.9           & \bfseries98.1            & 76.4           & 84.6 \\
    \ w/ SCP   &  \bfseries87.4           & \bfseries83.2          & 96.6           & \bfseries78.3         &  \bfseries86.4\\
    \hline
    ER \cite{zhao2020domain}     &  87.5            & 79.3           & \bfseries98.3            & 76.3           & 85.3 \\
    \ w/ SCP   &  \bfseries87.5           & \bfseries81.5          & 97.8           & \bfseries80.2         &  \bfseries86.8\\
    \hline
    MDA (ours)      &  \bfseries87.3            & 81.5           & \bfseries96.8            & 82.1           & 86.9 \\
    \ w/ SCP   &  87.1           & \bfseries85.2          & 96.7           & \bfseries83.5         &  \bfseries88.1\\
    \bottomrule
    \end{tabular}}
    \label{benefit}
    \end{table}

    \begin{figure*}[t!]
        \centering
        
        \subfloat[DANN]{\includegraphics[width=1.6in]{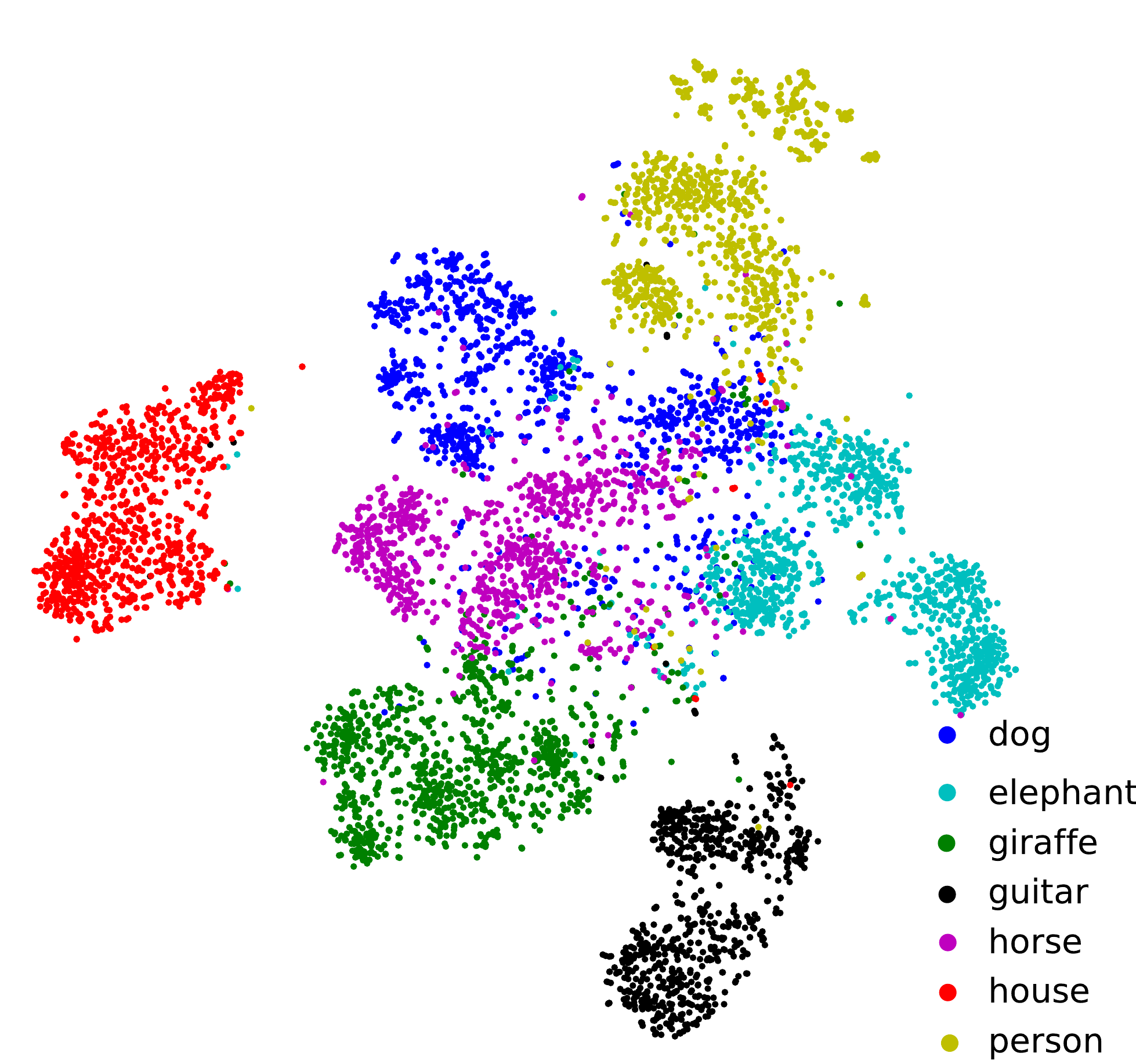}}\label{DANN}
        \hspace{0.3cm}
        \subfloat[CDANN]{\includegraphics[width=1.6in]{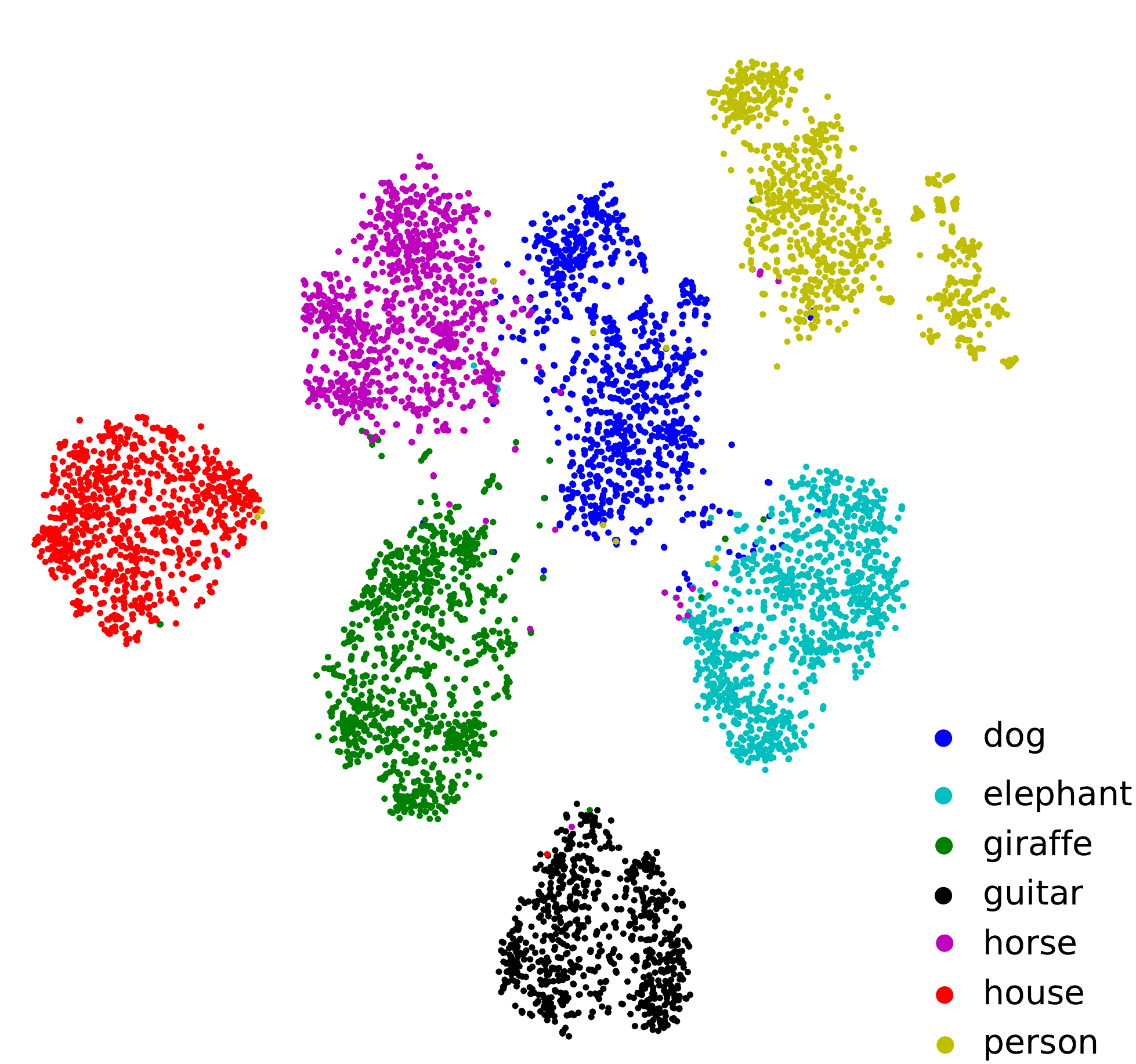}}\label{CDANN}
        \hspace{0.3cm}
        \subfloat[ER]{\includegraphics[width=1.6in]{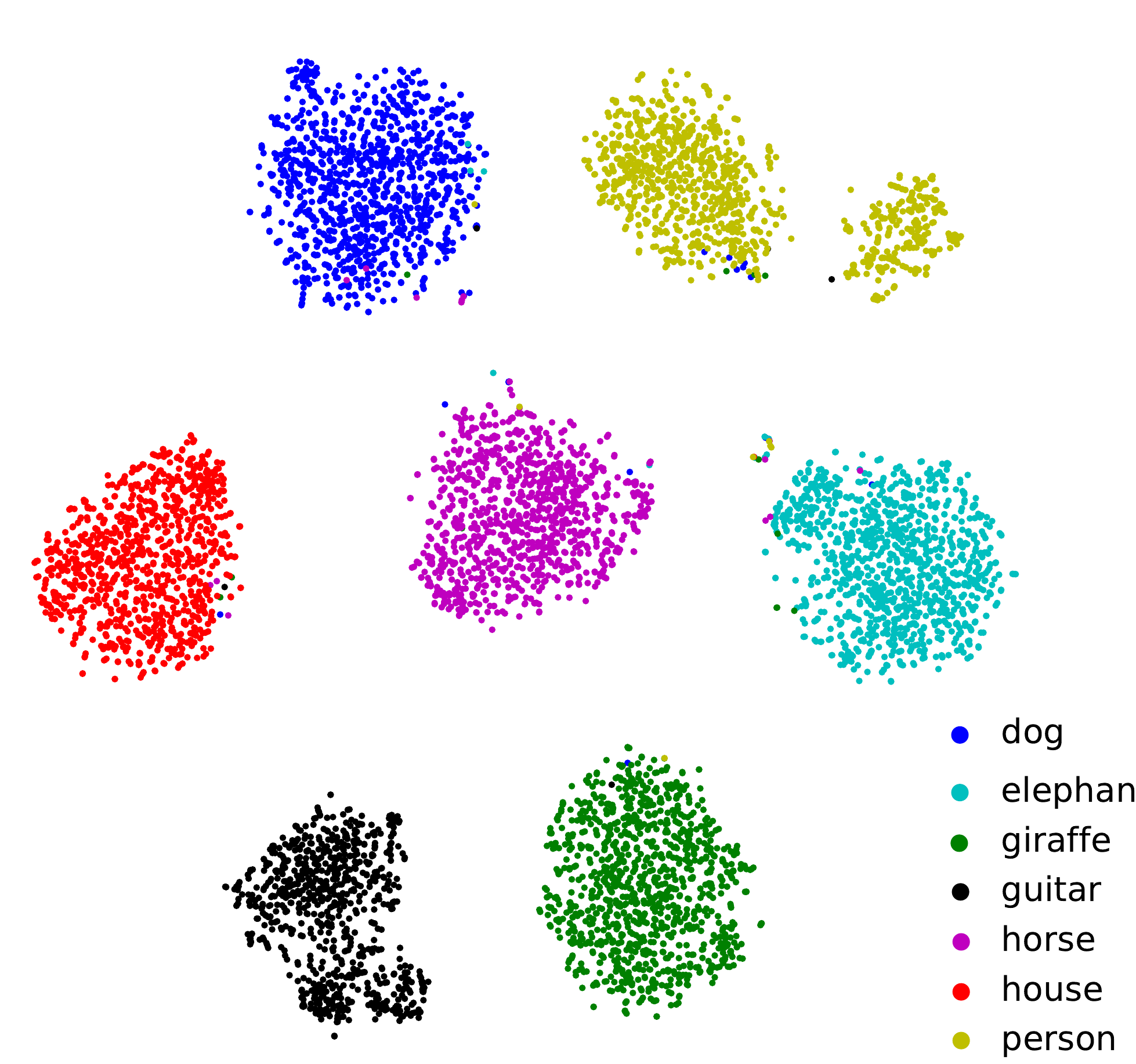}}\label{ER}
        \hspace{0.3cm}
        \subfloat[MDA (ours)]{\includegraphics[width=1.6in]{Figure/new_jdm.pdf}}\label{SAACF}
        \centering
        \caption{Visualizations with t-SNE embeddings~\cite{van2008visualizing} illustrating various classes' representations 
        produced by (a) DANN~\cite{ganin2016domainadversarial}, (b) CDANN~\cite{li2018deep}, 
        (c) ER~\cite{zhao2020domain}, and (d)~MDA (ours), respectively. MDA demonstrates the superior clustering effect. Zoom in for details.}
        \label{t-sne}
    
    \end{figure*}

\noindent
\textbf{Comparison with Dropout.}
Our proposed SCP endeavors to prune inconsistent channels across domains, aiming to eliminate spurious correlations rather than merely mitigating the interdependence among neurons akin to Dropout.
To substantiate the effectiveness of SCP in eliminating spurious correlations, experiments were undertaken on various benchmarks utilizing both Dropout~\cite{srivastava2014dropout} and our novel SCP. The outcomes, as illustrated in TABLE~\ref{dropout}, reveal a notable improvement in the generalization performance of SCP by 2.1\%, 0.1\%, 6.0\%, 2.3\%, and 3.2\% on PACS, VLCS, OfficeHome, TerraIncognita and DomainNet, respectively. These consistent findings consistently affirm the superior ability of SCP to eliminate spurious features, thereby enhancing feature disciminability.

\noindent\textbf{Feature Visualization with SCP.}
To visually evaluate the efficacy of the proposed SCP, we visualize the acquired features with t-SNE embeddings~\cite{van2008visualizing}. Specifically, we conduct experiments on PACS and take each domain as the target domain in turn. The clustering results, based on the features generated by ERM and SCP-based ERM, are depicted in Fig.~\ref{ERM_TSNE} and Fig.~\ref{MASK_TSNE}, respectively. 
Evidently, the introduced SCP module consistently improves the representation learning across all scenarios. The observed high inter-class separation and intra-class compactness in all scenarios further affirm the superiority of SCP for enhancing feature discriminability.

\noindent\textbf{Integration into Distribution Matching Models.}
To further assess the effectiveness of SCP in enhancing feature discriminability, we integrate SCP with distribution matching models, namely DANN~\cite{ganin2016domainadversarial}, CDANN~\cite{li2018deep}, ER~\cite{zhao2020domain}, and the proposed MDA. In our integration, we refrain from fine-tuning and maintain the same hyperparameters of these distribution matching techniques. This restraint underscores the robustness of SCP's effectiveness. The corresponding generalization performances are presented in TABLE~\ref{benefit}. The results indicate that SCP significantly enhances the generalization performance of these distribution matching methods, particularly in hard-to-transfer domains. These observations reinforce the superior capabilities of SCP in improving feature discriminability.

    \begin{figure*}[t!]
        \centering
        \subfloat[DANN]{\includegraphics[width=1.7in]{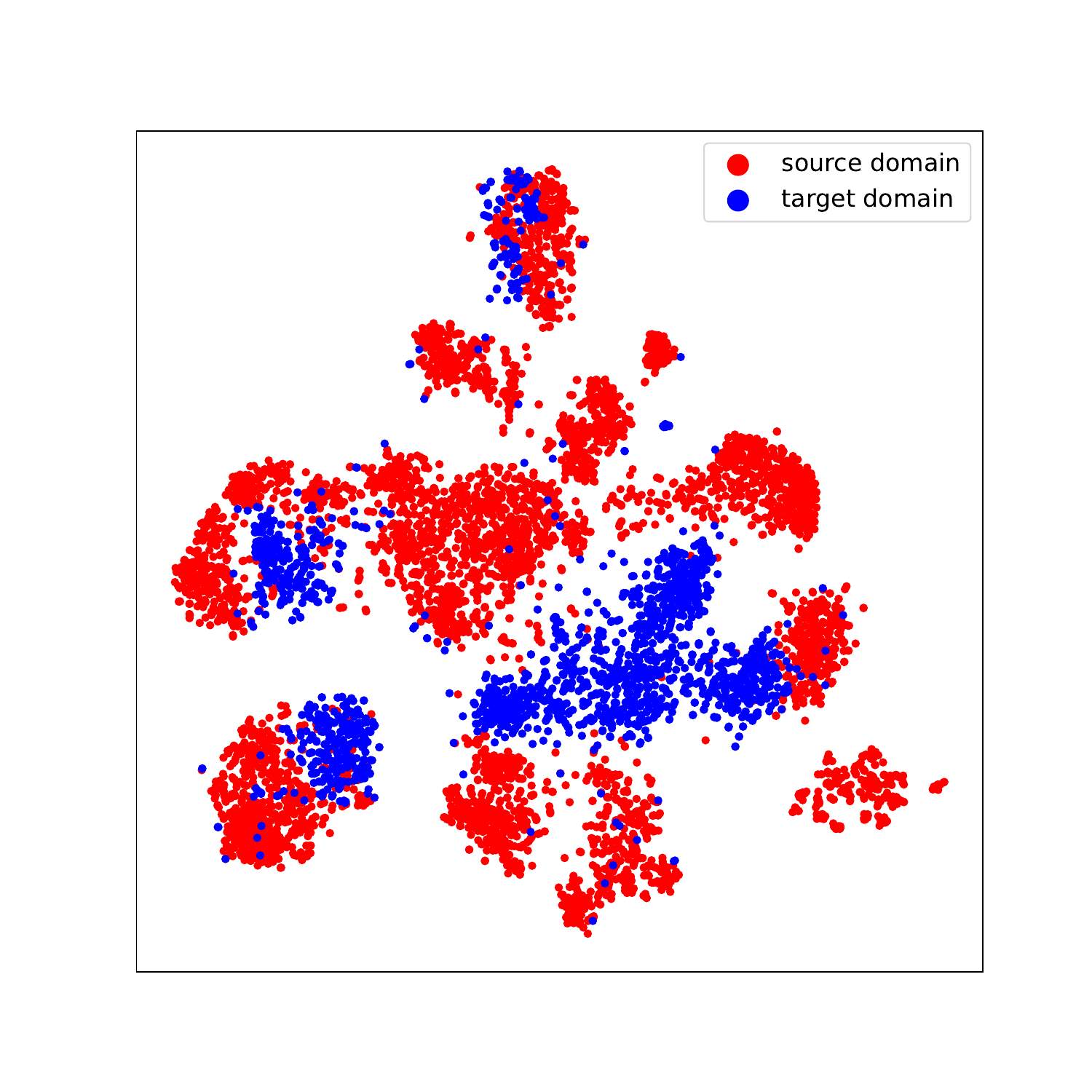}}\label{DANNS_T}
        \subfloat[CDANN]{\includegraphics[width=1.7in]{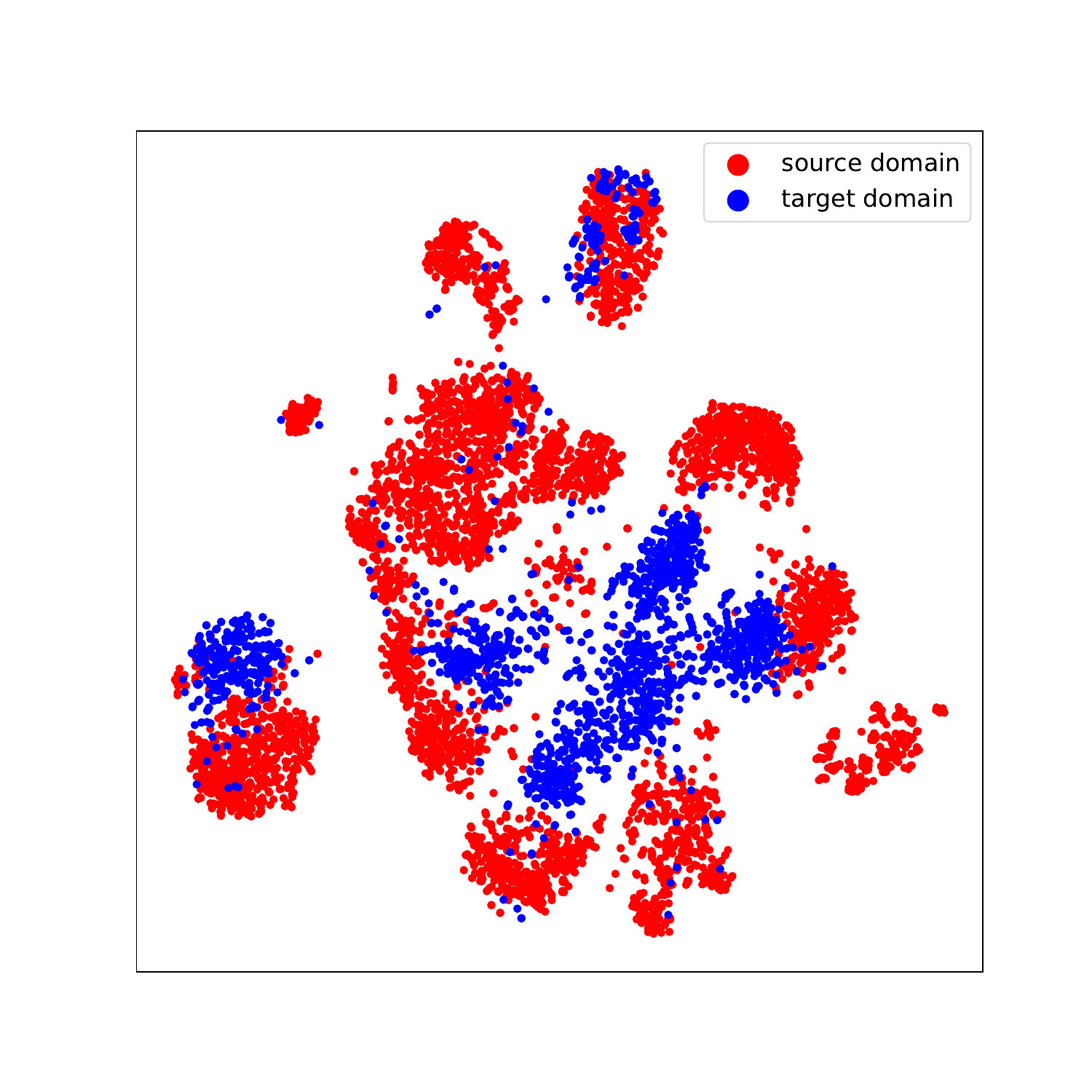}}\label{CDANNS_T}
        \subfloat[ER]{\includegraphics[width=1.7in]{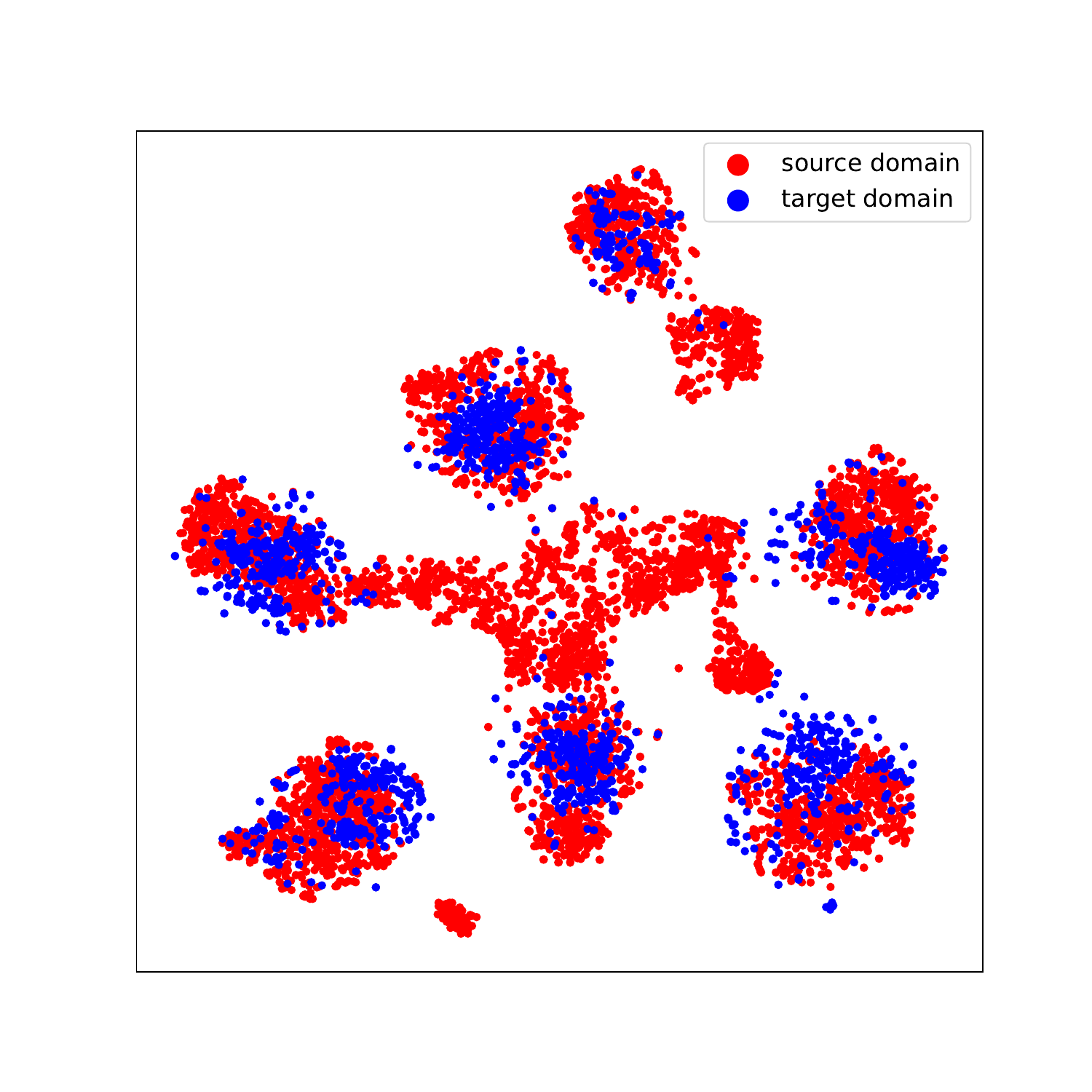}}\label{ERS_T}
        \subfloat[MDA (ours)]{\includegraphics[width=1.7in]{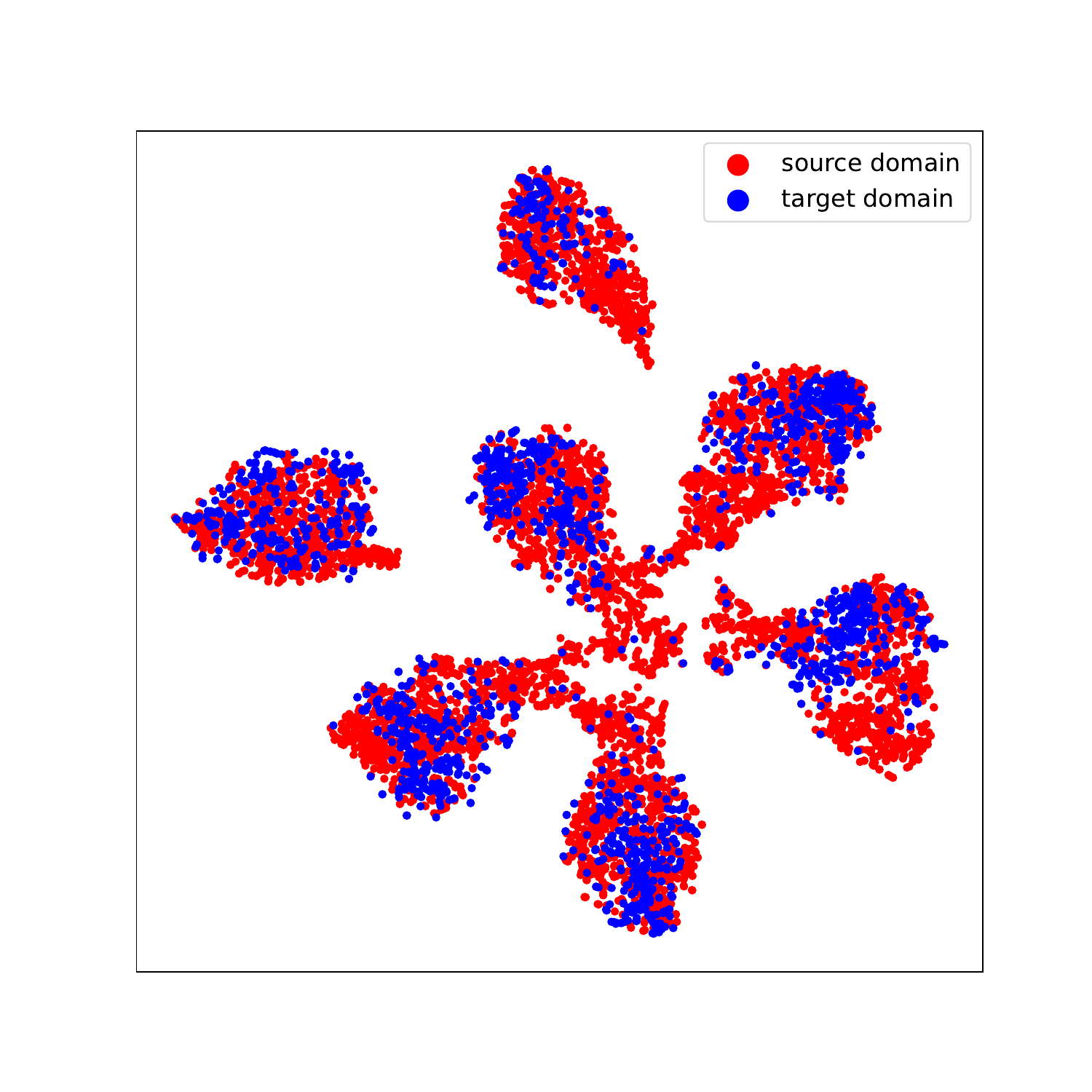}}\label{JDMS_T}
        \centering
        \caption{Visualizations with t-SNE embeddings~\cite{van2008visualizing} for the representations 
        across source domain and target domain learned by (a) DANN~\cite{ganin2016domainadversarial}, 
        (b) CDANN~\cite{li2018deep}, (c) ER~\cite{zhao2020domain}, and (d) MDA, respectively.
        The distribution discrepancy of MDA is the smallest.}
        \label{ST_stne}
    
    \end{figure*}

\begin{figure*}[t!]
  \centering
  \subfloat[$\mathcal{A}$-distance of features \label{adis}]{\includegraphics[width=0.25\linewidth, height=1.5in]{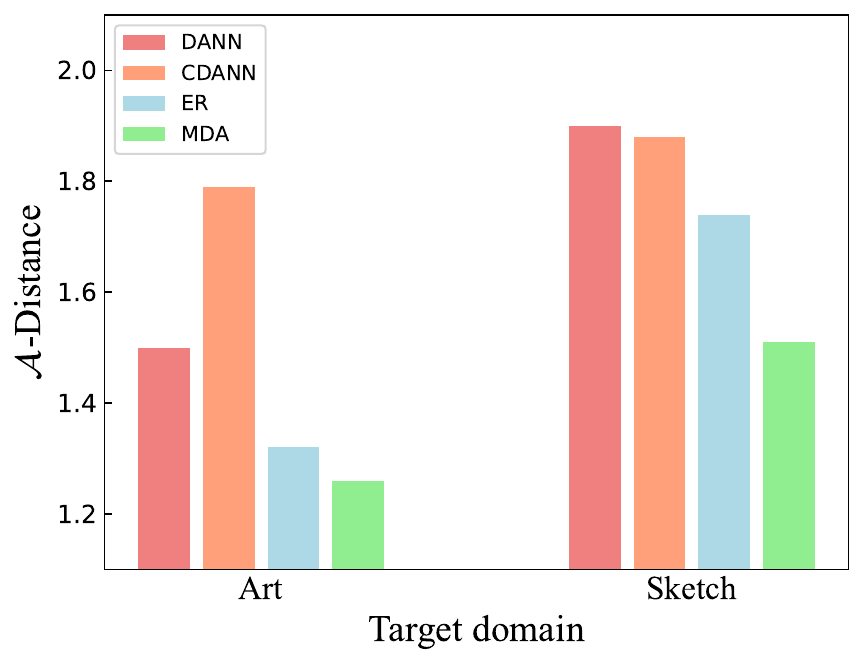}}
  \hspace{2mm}
  \subfloat[Effect of $m$ in SCP \label{m}]{\includegraphics[width=0.22\linewidth, height=1.5in]{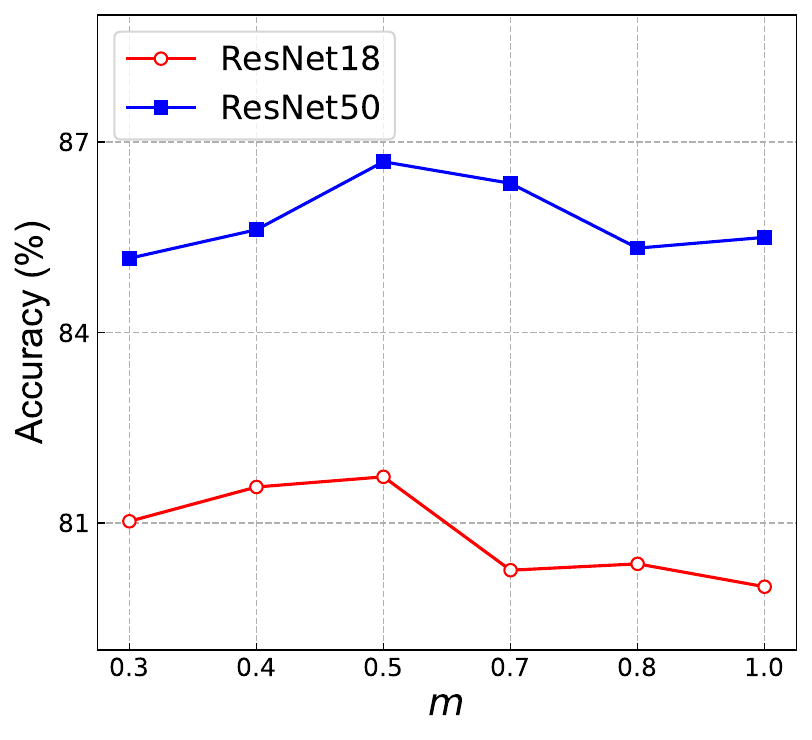}}
  \hspace{2mm}
  \subfloat[Effect of $\alpha$ for MDA \label{alpha}]{\includegraphics[width=0.22\linewidth, height=1.5in]{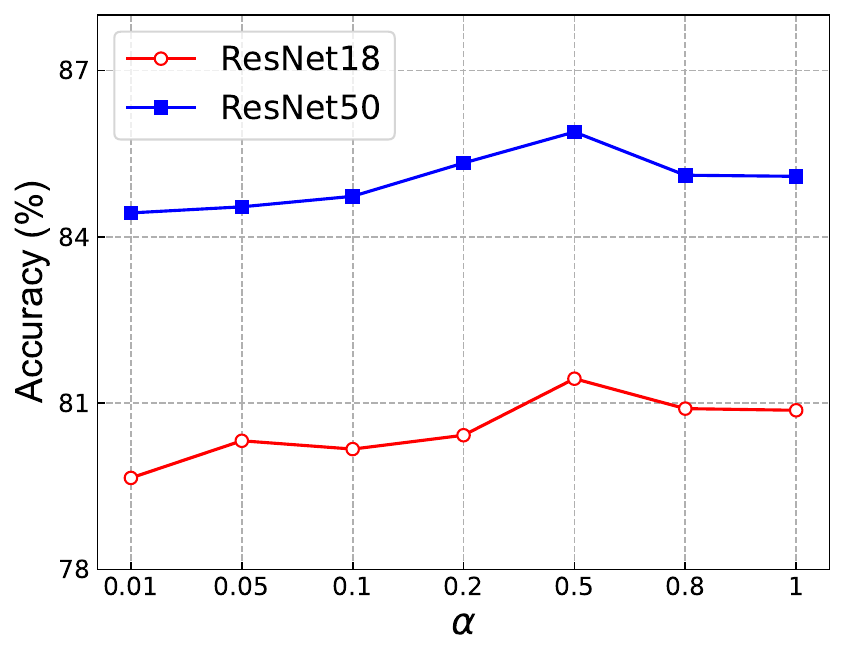}}
  \hspace{2mm}
  \subfloat[Effect of $\beta$ for SCP \label{beta}]{\includegraphics[width=0.17\linewidth, height=1.5in]{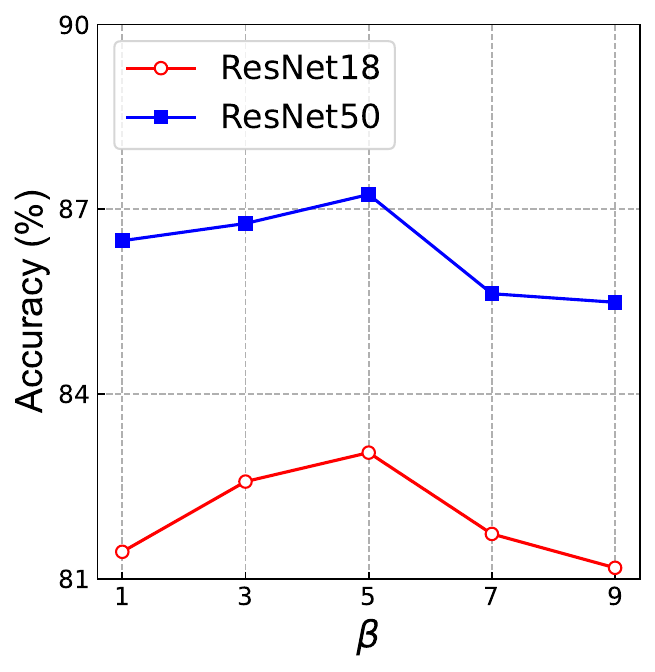}}
   \caption{(a) $\mathcal{A}$-distance of features learned by DANN, CDANN, ER, and the proposed MDA, respectively. Lower is better. (b) Generalization performance of SCP w.r.t. the quantile $m$. (c) Generalization performance of MDA w.r.t. the hyperparameter $\alpha$. (d) Performance of SCP w.r.t. the coefficient $\beta$.   \label{weighting}}

\end{figure*}

\begin{table}
\caption{Comparison of distribution matching models with distinct backbones on PACS. Bold indicates the best performance.}
    \centering
    \setlength{\tabcolsep}{3pt}
    {
    \begin{tabular}{l|cccc|c}
    \toprule
    Method & Art & Cartoon & Photo & Sketch & Avg.($\uparrow$) \\
    \midrule
    \multicolumn{6}{c}{ResNet-18}\\
    \hline
    DANN \cite{ganin2016domainadversarial}  & 77.4 & 66.8 & 96.6  & 69.8 & 77.7   \\
    CDANN \cite{li2018deep}       & 75.2  & 74.0 & 95.7  & 72.9 & 79.4 \\
    ER \cite{zhao2020domain}        & 80.7  & 76.4 & \bfseries96.7  & 71.8 & 81.4 \\
    MDA (ours)      & \bfseries81.0   & \bfseries78.5 & 95.5  & \bfseries77.4 & \bfseries83.1\\
    \hline
    \multicolumn{6}{c}{ResNet-50}\\
    \hline
    DANN \cite{ganin2016domainadversarial}  & 86.4  & 77.4 & 97.3  & 73.5 & 83.7   \\
    CDANN \cite{li2018deep}       & 85.0  & 78.9 & 98.1  & 76.4 & 84.6 \\
    ER \cite{zhao2020domain}        & 87.5  & 79.3 & \bfseries98.3  & 76.3 & 85.3 \\
    MDA (ours)      & \bfseries87.3    & \bfseries81.5 & 96.8  & \bfseries82.1 & \bfseries86.9 \\
    \bottomrule
 
    \end{tabular}}
        \vspace{-2mm}
    \label{PACS_res50}
    \end{table}   

\noindent\textbf{Comparison with Existing Distribution Matching Models.}
To demonstrate the superiority of the proposed MDA over existing distribution matching models, we conduct experiments on PACS using DANN~\cite{ganin2016domainadversarial}, 
CDANN~\cite{li2018deep}, ER~\cite{zhao2020domain}, and MDA, respectively.
The results, obtained with various backbones, as shown in TABLE~\ref{PACS_res50}, consistently affirm MDA's superiority in terms of generalization performance. Particularly noteworthy is MDA's substantial enhancement in hard-to-transfer domains such as `Cartoon' and `Sketch', where other distribution matching methods struggle significantly.

\noindent\textbf{Feature Visualization with MDA.}
To visually illustrate the impact of MDA, we employ t-SNE embeddings to visualize the acquired features. Specifically, we take experiments on PACS, with `Photo' as the target domain. 
Fig.~\ref{t-sne} depicts the clustering results based on the features generated by DANN~\cite{ganin2016domainadversarial}, CDANN~\cite{li2018deep}, ER~\cite{zhao2020domain}, and 
the proposed MDA, respectively. Notably, MDA demonstrates superior representations characterized by high inter-class separation and intra-class compactness, particularly when compared to DANN and CDANN.  While the improvement over ER may not be pronounced, notably observations can still be made. Firstly, the clustering of `guitar'~(black) and `person'~(yellow) is noticeably more compact with MDA. Secondly, the peripheries of class clusters exhibit greater dispersion with ER, highlighting the better representation achieved by the proposed MDA.

\noindent\textbf{Distribution Discrepancy.}
To assess the impact of MDA on domain invariance, we employ the $\mathcal{A}$-distance as a metric to gauge the distribution divergence across domains~\cite{ben2010theory}. 
A proxy $\mathcal{A}$-distance can be defined as $d_{\mathcal{A}} = 2(1-2\sigma)$, 
where $\sigma$ represents the error of a binary discriminator in correctly distinguishing between source and target domain samples. In our case, the discriminator is implemented as a neural network with a single FC layer. As shown in Fig.~\ref{adis}, MDA achieves a representation space characterized by a smaller domain discrepancy when compared to other distribution matching approaches. We further present t-SNE embeddings~\cite{van2008visualizing} of the acquired features from source domains and target domain in Fig.~\ref{ST_stne}. Notably, it is evident that the distributions of the acquired features across the source domain and target domain, when MDA is employed, exhibit a closer alignment. This observation substantiates MDA's capacity to learn features with enhanced generalizability, surpassing the capabilities of DANN~\cite{ganin2016domainadversarial}, CDANN~\cite{li2018deep} and ER~\cite{zhao2020domain}.

\noindent
\textbf{Hyperparameter Influence.}
We explore the influence of the quantile parameter $m$ in SCP, trade-off parameter $\alpha$ for MDA and $\beta$ for SCP on the generalization performance. Fig.~\ref{m}, Fig.~\ref{alpha} and Fig.~\ref{beta} illustrate the generalization accuracies of SCP and MDA, utilizing two different pre-trained backbones, ResNet-18 and ResNet-50. We systematically vary $m$ within the range $\{0.3, 0.4, 0.5, 0.7, 0.8, 1.0\}$, $\alpha$ within the range $\{0.01, 0.05, 0.1, 0.2, 0.5, 0.8, 1.0\}$, and $\beta$ within the range $\{1, 3, 5, 7, 9\}$. The consistent trend of initially increasing and subsequently decreasing generalization performance observed in both SCP and MDA underscores their vital role in enhancing generalization performance. For the selected trade-off values in the overall objectives, we consistently set $\alpha$ to 1.0 throughout all experiments to ensure the effectiveness of MDA, while employing $\beta$ values of 3 and 5 consistently across all experiments, thereby underscoring the robustness of our proposed modules. Subsequently, we select the optimal experimental outcome from these options of $\beta$.

\section{Conclusion}
\label{Con}

In this paper, we present a novel perspective on DG by concurrently enhancing feature generalizability while elevating feature discriminability. We introduce a novel framework for domain-invariant representation learning, named Discriminative Microscopic Distribution Alignment~(DMDA), comprising two key modules: Selective Channel Pruning~(CAP) and Micro-level Distribution Alignment~(MDA). SCP is designed to enhance feature discriminability by eliminating spurious correlations within neural networks. It achieves this by selectively pruning unstable channels. On the other hand, MDA attempts to excavate sufficient generalizable features while accommodating within-class variations through a micro-level alignment manner. Experiments on four benchmark datasets demonstrate the efficacy of DMDA in learning an increased representation space characterized by superior discriminability and generalizability, thereby leading to improved generalization performance. Our proposed method sheds light on the significance of stable factors and micro-level alignment in domain-invariant representation learning for DG.

{
\bibliographystyle{IEEEtran}
\bibliography{egbib}
}

\begin{IEEEbiography}[{\includegraphics[width=1in,height=1.25in,clip,keepaspectratio]{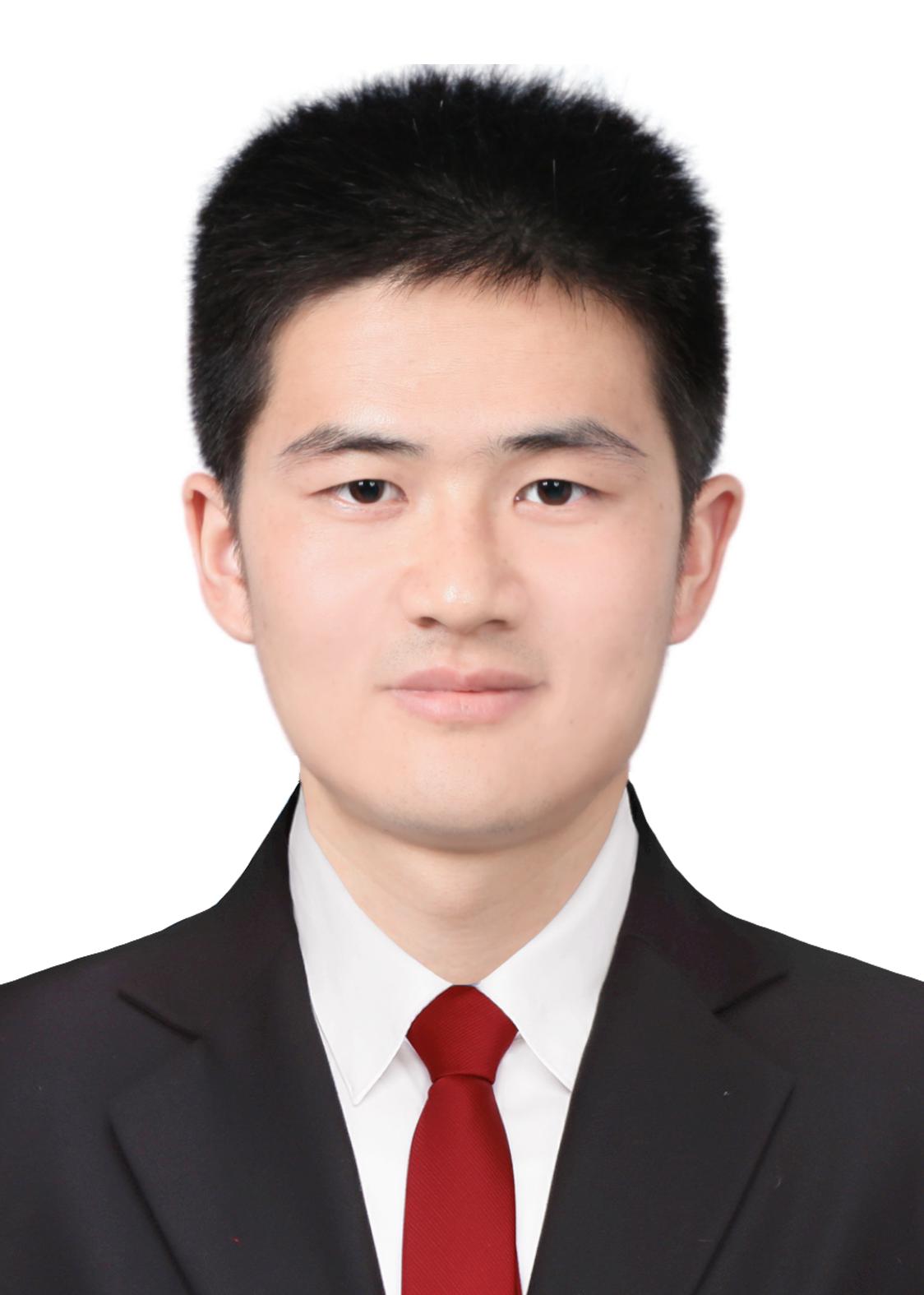}}]{Shaocong Long}
is currently pursuing his Ph.D. degree in the Department of Computer Science and Engineering, Shanghai Jiao Tong University. Before that, he received a B.E. degree in Nankai University in 2019. His current research interests focus on domain generalization, causal inference.
\end{IEEEbiography}

\vspace{-0.1cm}

\begin{IEEEbiography}[{\includegraphics[width=1in,height=1.25in,clip,keepaspectratio]{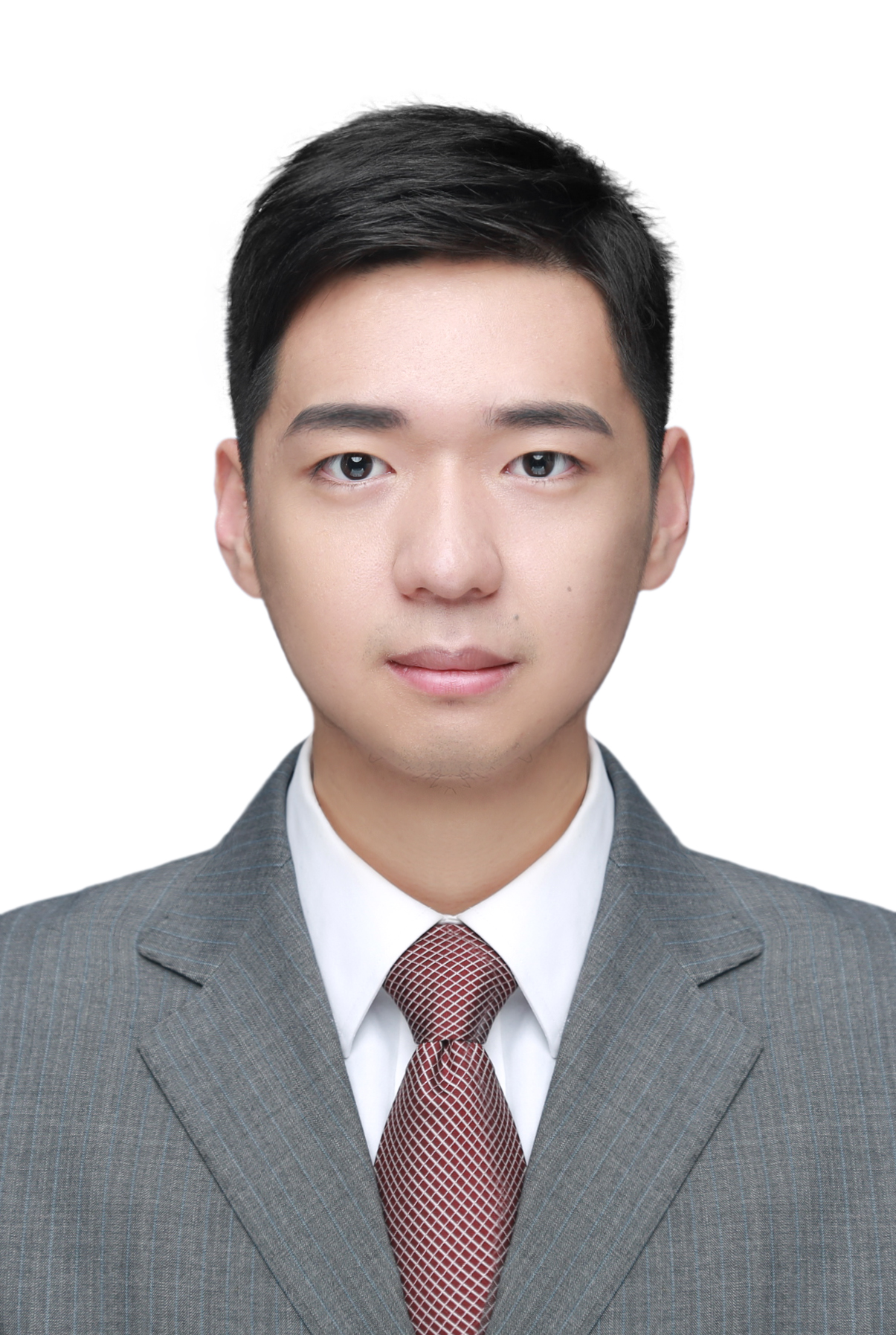}}]{Qianyu Zhou}
is currently pursuing his Ph.D. degree in the Department of Computer Science and Engineering, Shanghai Jiao Tong University.   Before that, he received a B.Sc. degree in Jilin University in 2019.   His current research interests focus on computer vision, scene understanding, transfer learning.  He serves regularly as a reviewer for IEEE TPAMI, IJCV, IEEE TIP, IEEE TCSVT, IEEE TMM, CVPR, ICCV, ECCV, AAAI, etc.
\end{IEEEbiography}

\vspace{-0.1cm}

\begin{IEEEbiography}[{\includegraphics[width=1in,height=1.25in,clip,keepaspectratio]{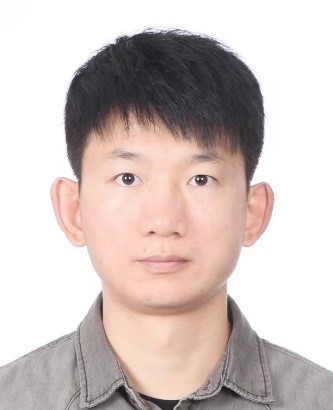}}]{Chenhao Ying}
received the B.E. degree in the Department of Communication Engineering, Xidian University, China, in 2016, and the Ph.D. degree in the Department of Computer Science and Engineering, Shanghai Jiao Tong University, China, in 2022. He is currently a research assistant professor in the Department of Computer Science and Engineering, Shanghai Jiao Tong University. His current research interests include mobile crowd sensing, blockchain and federated learning.
\end{IEEEbiography}

\vspace{-0.1cm}

\begin{IEEEbiography}[{\includegraphics[width=1in,height=1.25in,clip,keepaspectratio]{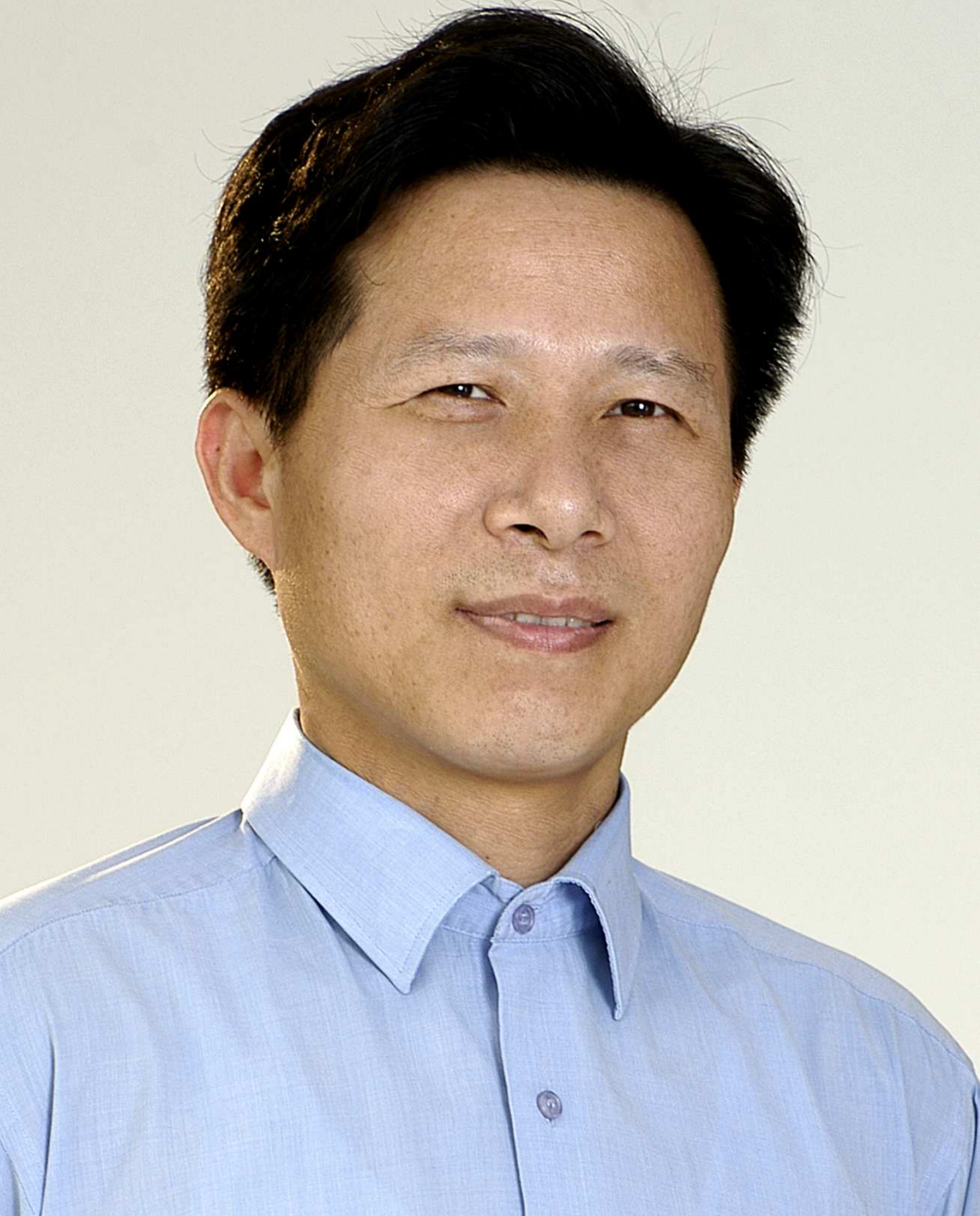}}]{Lizhuang Ma} received his B.S. and Ph.D. degrees from the Zhejiang University, China in 1985 and 1991, respectively. He is now a Distinguished Professor, Ph.D. Tutor, and the Head of the Digital Media and Computer Vision Laboratory at the Department of Computer Science
and Engineering, Shanghai Jiao Tong University, China. He was a Visiting Professor at the
Frounhofer IGD, Darmstadt, Germany in 1998,
and was a Visiting Professor at the Center for
Advanced Media Technology, Nanyang Technological University, Singapore from 1999 to 2000. He published more than 400 academic research papers in both domestic and international
journals. His research interests include computer-aided geometric design, computer graphics, computer vision, \emph{etc}. 
\end{IEEEbiography}

\vspace{-0.1cm}

\begin{IEEEbiography}[{\includegraphics[width=1in,height=1.25in,clip,keepaspectratio]{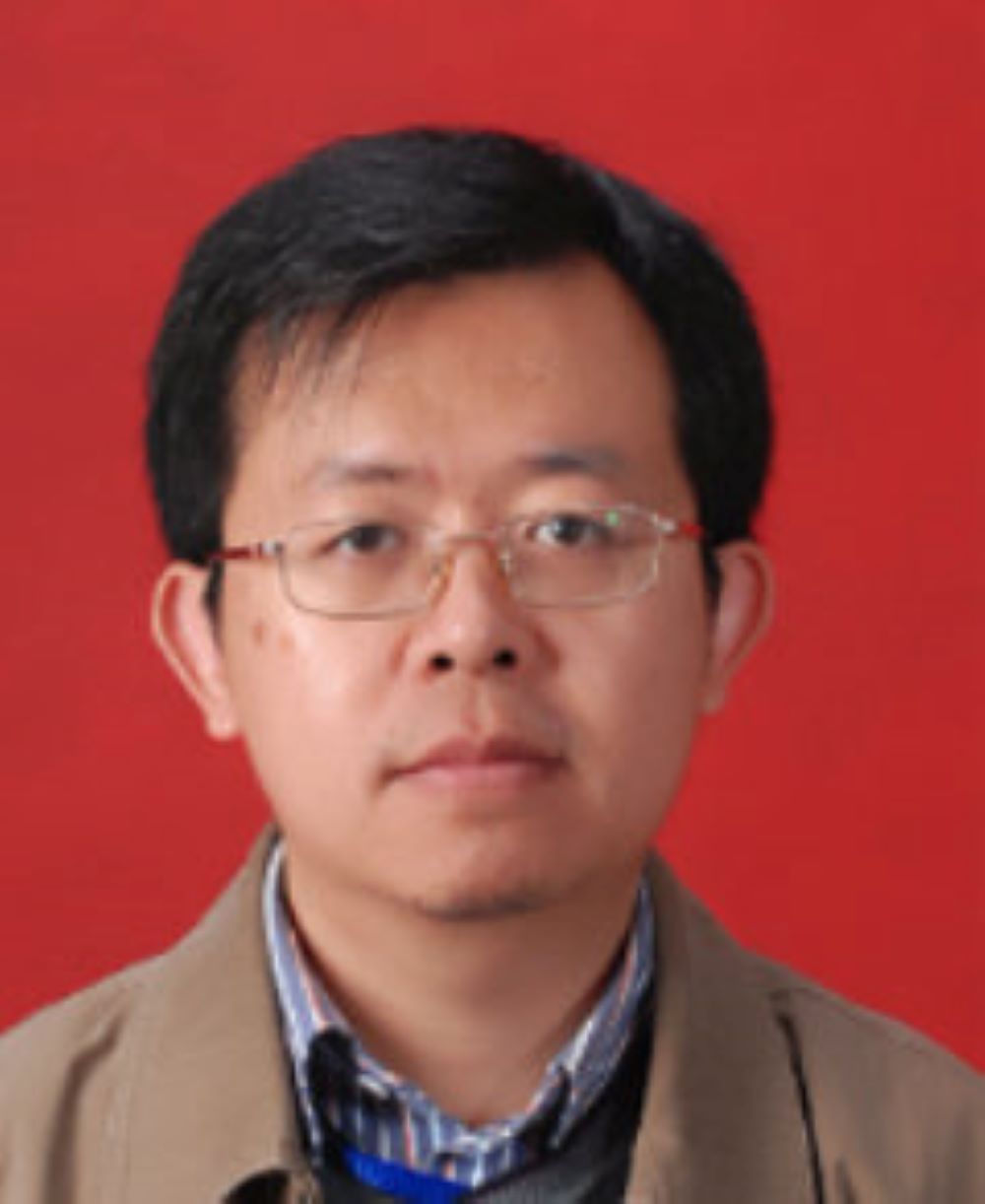}}]{Yuan Luo}
received the B.S. degree in applied
mathematics and the M.S. and Ph.D. degrees
in probability statistics from Nankai University,
Tianjin, China, in 1993, 1996, and 1999, respectively.
From July 1999 to April 2001, he held a
post-doctoral position with the Institute of Systems
Science, Chinese Academy of Sciences,
Beijing, China. From May 2001 to April 2003,
he held another post-doctoral position with the
Institute for Experimental Mathematics, University
of Duisburg-Essen, Essen, Germany. Since
June 2003, he has been with the Department of Computer Science and
Engineering, Shanghai Jiao Tong University, Shanghai, China. Since
2006, he has been a Full Professor and the Vice Dean of the department
(from 2016 to 2018 and since 2021). His current research interests
include coding theory, information theory, and big data analysis.
\end{IEEEbiography}

\appendix
\section{Proofs}

\subsection{Proof for Proposition 1}

\begin{proof}
For the fixed semantics experts $\{E_i\}_{i = 1}^M$ and feature extractor $f$, the minimax game in Eq.~\eqref{minmax} in the main text reduces to the following maximizing problem:
\begin{equation}
\begin{split}
   & D^{\ast} = \{D^{\ast}_1(\Phi^{\prime}, S^{\prime}), D^{\ast}_2(\Phi^{\prime}, S^{\prime}), \cdots, D^{\ast}_M(\Phi^{\prime}, S^{\prime})\} \\
    = &\arg \max_{D} \sum_{i = 1}^M \sum_{x, y}P^S_i(x, y)\log D_i(\Phi^{\prime}, S^{\prime}), \\
    = &\arg \max_{D} \sum_{i = 1}^M \sum_{\Phi^{\prime}, S^{\prime}}P^S_i(\Phi^{\prime}, S^{\prime})\log D_i(\Phi^{\prime}, S^{\prime}), \\
& s.t.\quad \sum_{i = 1}^M D_i(\Phi^{\prime}, S^{\prime}) = 1. \label{proof1}
\end{split}
\end{equation}%
We address this problem by maximizing the function point-wisely and then employing the Lagrangian multiplier:
\begin{equation}
\begin{split}
    &D^{\ast} = \{D^{\ast}_1(\Phi^{\prime}, S^{\prime}), D^{\ast}_2(\Phi^{\prime}, S^{\prime}), \cdots, D^{\ast}_M(\Phi^{\prime}, S^{\prime})\}\\
    =& \arg \max_{D} (\sum_{i = 1}^M P^S_i(\Phi^{\prime}, S^{\prime})\log D_i(\Phi^{\prime}, S^{\prime}) \\
    &+\lambda(\sum_{i = 1}^M D_i(\Phi^{\prime}, S^{\prime}) -1)). \label{proof11}
\end{split}
\end{equation}%

Setting the derivative of Eq.~\eqref{proof11} w.r.t. $D_i(\Phi^{\prime}, S^{\prime})$ to zero and combining the constraint $\sum_{i = 1}^M D_i(\Phi^{\prime}, S^{\prime}) = 1$, we can obtain the following conclusion:
\begin{equation}
\begin{split}
 \hspace{-2.8mm} \lambda = -\sum_{i = 1}^M  P^S_i(\Phi^{\prime}, S^{\prime}),
  D^{\ast}_i(\Phi^{\prime}, S^{\prime}) = \frac{P^S_i(\Phi^{\prime}, S^{\prime})}{\sum^M_{j = 1}P^S_j(\Phi^{\prime}, S^{\prime})}. \label{proof12}
\end{split}
\end{equation}%

\end{proof}

\subsection{Proof for Theory 1}
\begin{proof}
If $D^*$ is the optimum for the inner maximization of Eq.~\eqref{minmax} in the main text, then the minimax game of Eq.~\eqref{minmax} in the main text reduces to the following minimizing problem:
\begin{equation}
\begin{split}
  &\hspace{-3.8mm}\min_{f}-H(D(\Phi, S))
  = \min_{f}\sum_{i = 1}^{M}\sum_{x, y}{P^S_i(x, y)}\log \frac{P^S_i(\Phi, S)}{\sum^M_{j = 1}P^S_j(\Phi, S)}\\
  =&\min_{f}\sum_{i = 1}^{M}\sum_{\Phi, S}{P^S_i(\Phi, S)}\log \frac{P^S_i(\Phi, S)}{\sum^M_{j = 1}P^S_j(\Phi, S)} \\
  &+
   M \log M - M \log M\\
  =& \min_{f}\sum_{i = 1}^{M}\sum_{\Phi, S}{P^S_i(\Phi, S)}\log \frac{P^S_i(\Phi, S)}{\frac{1}{M}\sum^M_{j = 1}P^S_j(\Phi, S)} + C\\
  =& \min_{f} \sum_{i = 1}^{M} KL(P^S_i(\Phi, S)||\frac{1}{M}\sum^M_{j = 1}P^S_j(\Phi, S)) + C\\
  =& \min_{f} M \cdot JSD(P_1^S(\Phi, S), P_2^S(\Phi, S), \cdots, P_M^S(\Phi, S)) + C,
\end{split}
\end{equation}%
where $C = -M\log M$ is a constant. Thus the minimax game in Eq.~\eqref{minmax} in the main text is equivalent to matching the semantic-based distributions in Eq.~\eqref{matching} in the main text, and the global optimum of Eq.~\eqref{minmax} in the main text can be attained if and only if $P_1^S(\Phi, S) = P_2^S(\Phi, S) = \cdots = P_M^S(\Phi, S)$.
\end{proof}

\vfill

\end{document}